\documentclass[numbered]{trbunofficial}
\usepackage{pdflscape}
\usepackage{graphicx}
\usepackage{subcaption}
\usepackage{float}  
\usepackage{booktabs}
\usepackage{multirow}
\usepackage{array}
\usepackage[table,xcdraw]{xcolor}
\usepackage{pdfpages}
\usepackage{gensymb}

\usepackage[hidelinks]{hyperref}

\AuthorHeaders{Sharma et. al.}
\title{Introducing the transitional autonomous vehicle lane-changing dataset: Empirical Experiments}

\author{%
  \textbf{Abhinav Sharma}\\
  Ph.D. Student\\
  Department of Civil, Construction and Environmental Engineering\\
  North Carolina State University, Raleigh, North Carolina, USA\\
  asharm63@ncsu.edu\\
  \hfill\break
  \textbf{Zijun He}\\
  Ph.D. Student\\
  Department of Civil, Construction and Environmental Engineering\\
  North Carolina State University, Raleigh, North Carolina, USA\\
  zhe8@ncsu.edu\\
  \hfill\break
  \textbf{Danjue Chen, Ph.D., Corresponding Author}\\
    Associate Professor\\
    Department of Civil, Construction and Environmental Engineering\\
    North Carolina State University, Raleigh, North Carolina, USA\\
    dchen33@ncsu.edu
  \hfill\break
}

\TotalWords{5937}

\begin{document}
\nolinenumbers
\maketitle

\section{Abstract}

Transitional autonomous vehicles (tAVs)—which operate beyond SAE Level 1–2 automation but short of full autonomy—are increasingly sharing the road with human‑driven vehicles (HDVs). As these systems interact during complex maneuvers such as lane changes, new patterns may emerge with implications for traffic stability and safety. Assessing these dynamics, particularly during mandatory lane changes, requires high‑resolution trajectory data, yet datasets capturing tAV lane‑changing (LC) behavior are scarce.

This study introduces the North Carolina Transitional Autonomous Vehicle Lane‑Changing (NC‑tALC) Dataset, a high‑fidelity trajectory resource designed to characterize tAV interactions during LC maneuvers. It comprises two controlled experimental series: (1) tAV LC Experiments, in which a tAV executes lane changes in the presence of adaptive‑cruise‑control (ACC)–equipped target vehicles, enabling analysis of LC execution; and (2) tAV Responding (Respd) Experiments, in which two tAVs act as followers and respond to cut‑in maneuvers initiated by another tAV, enabling analysis of follower response dynamics. The dataset contains 152 trials (72 LC, 80 Respd) sampled at 20 Hz with centimeter‑level RTK-GPS accuracy. The NC‑tALC dataset provides a rigorous empirical foundation enabling evaluation of tAV decision‑making and interaction dynamics in controlled mandatory LC scenario.


\hfill\break%
\noindent\textit{Keywords}: Mandatory Lane Changing, Empirical Experiments, Autonomous Vehicles, Driving Behavior, Trajectory dataset
\newpage

\newcommand{\figTestSite}{\begin{figure}[H]
  \centering
  \begin{subfigure}[t]{0.48\linewidth}
    \centering
    \includegraphics[width=\linewidth, height = 6 cm]{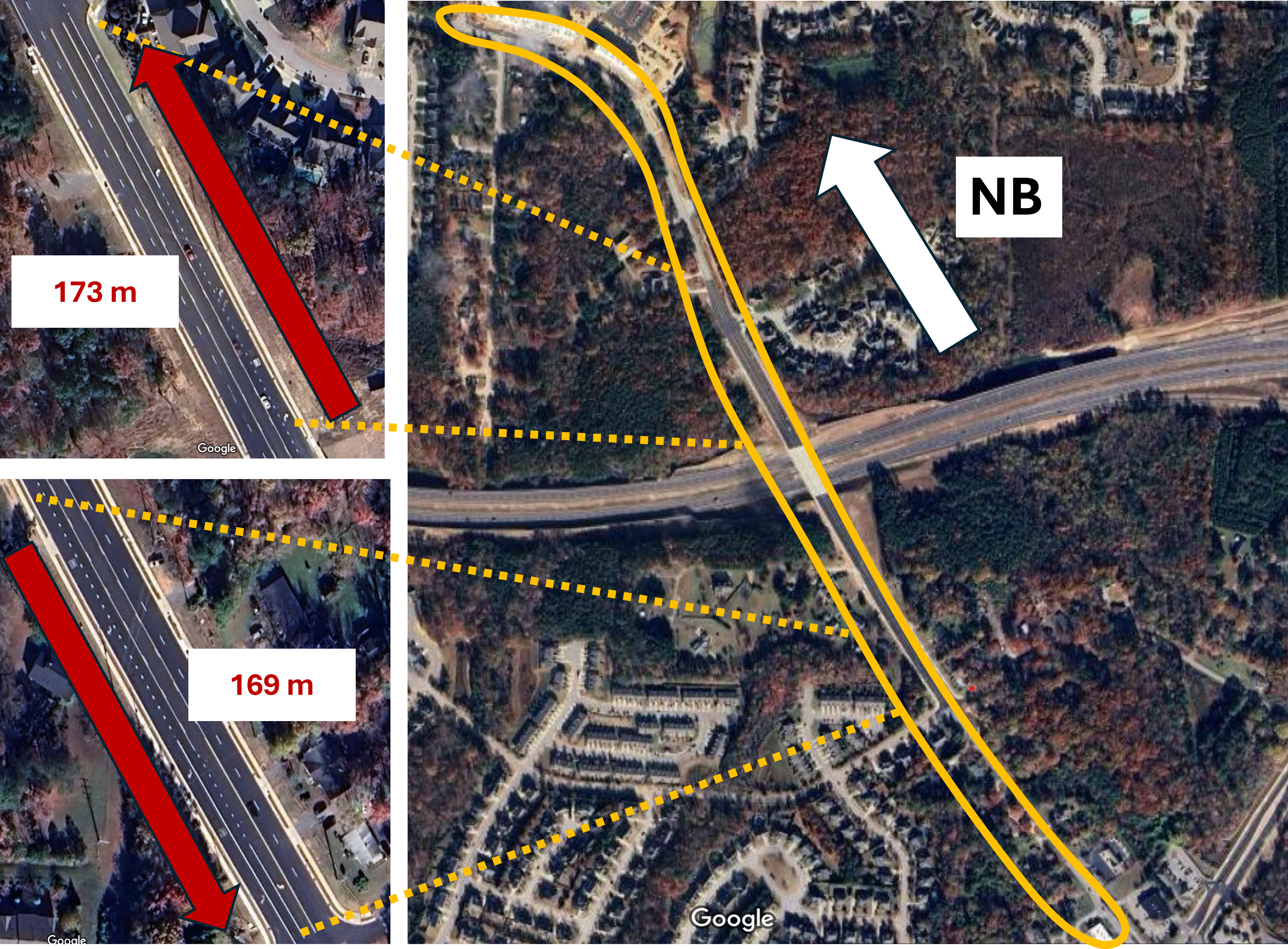}
    \caption{Test segment in NB and SB Direction; Google Maps (35.674718, -78.804330)}
    \label{fig:left}
  \end{subfigure}\hfill
  \begin{subfigure}[t]{0.48\linewidth}
    \centering
    \includegraphics[width=\linewidth, height = 6 cm]{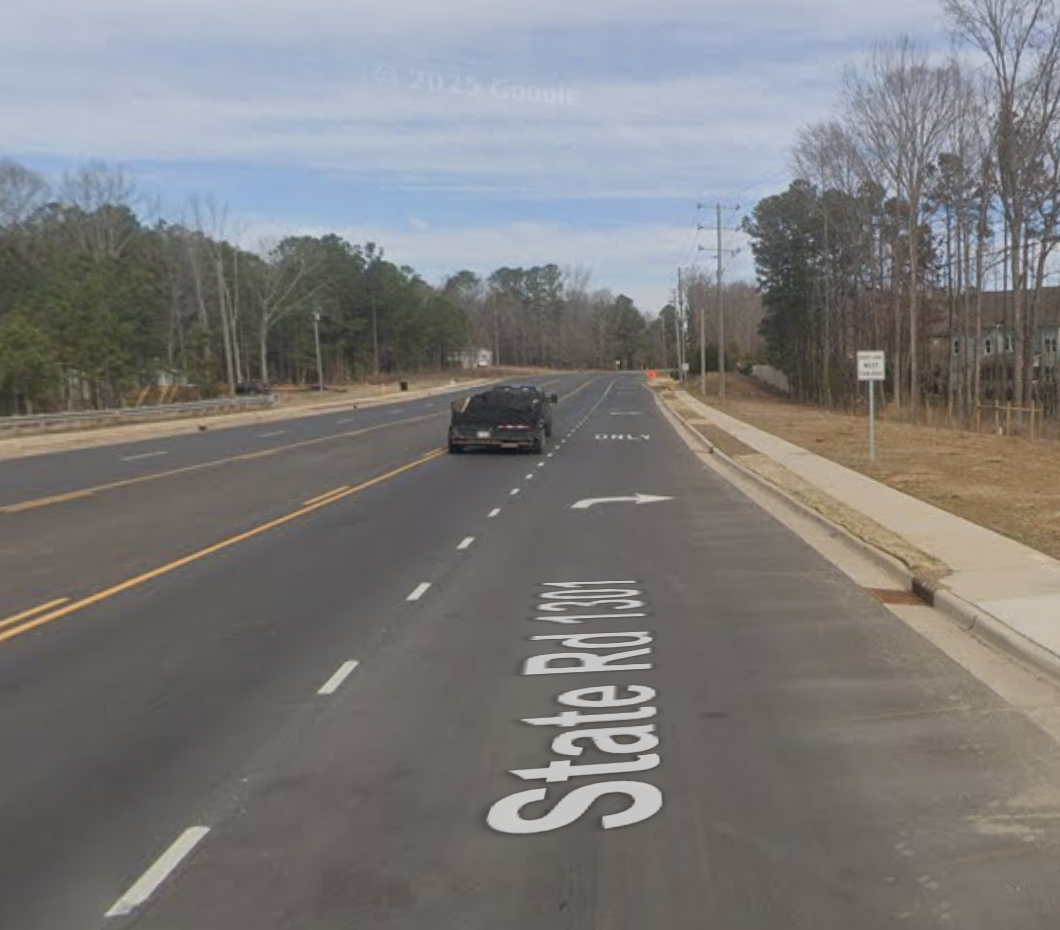}
    \caption{Right-turn lane in NB Direction}
    \label{fig:right}
  \end{subfigure}
  \caption{Experiment site in Apex, North Carolina, USA}
  \label{fig:Test Site}
\end{figure}}

\newcommand{\SiteLayout}{
\begin{figure}[H]
    \centering
    \includegraphics[width=\linewidth]{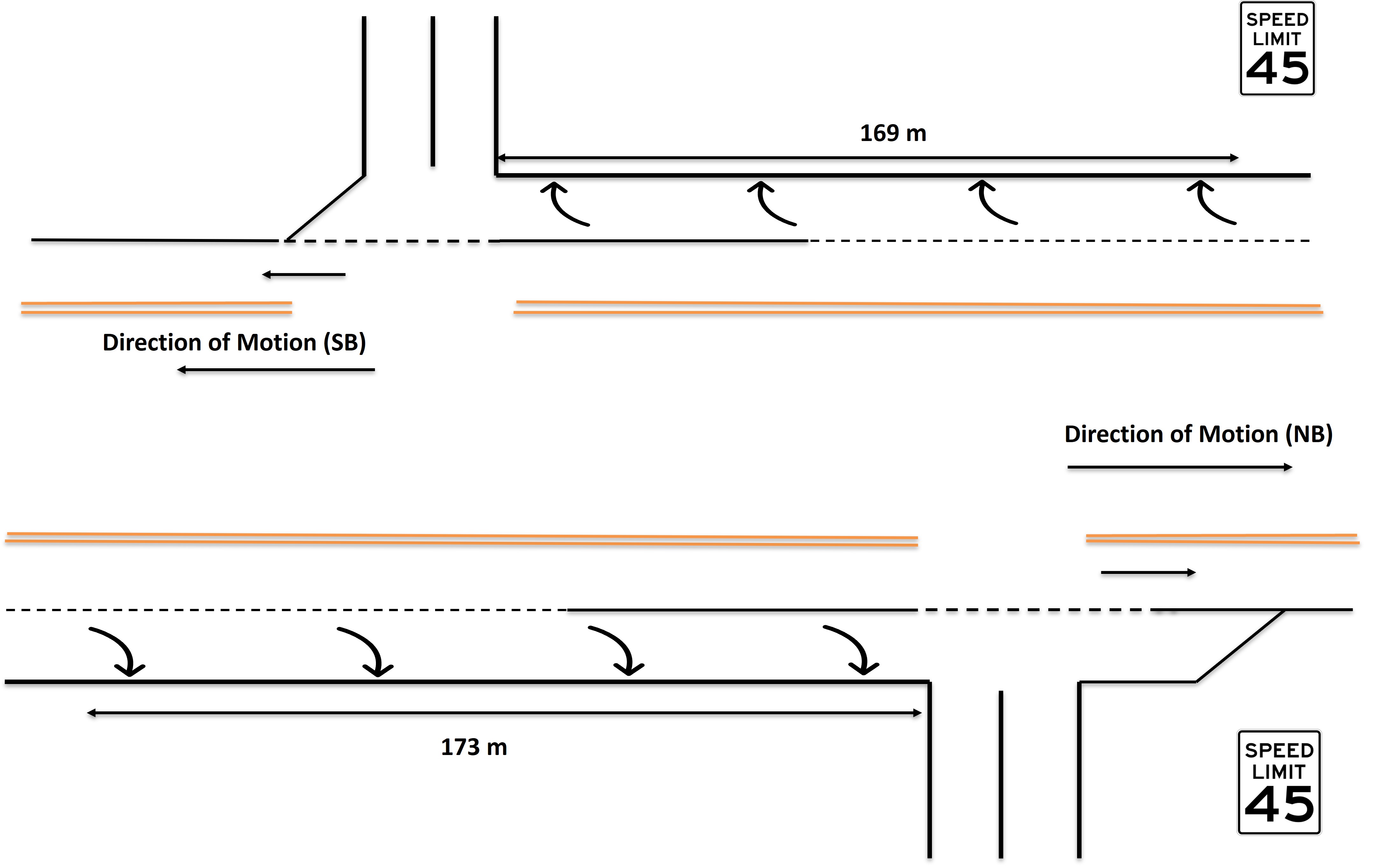}
    \caption{Geometric layout}
    \label{fig:Site Layout}
\end{figure}}

\newcommand{\figVehicleConfig}{
\begin{figure}[H]
    \centering
    \includegraphics[width=\linewidth]{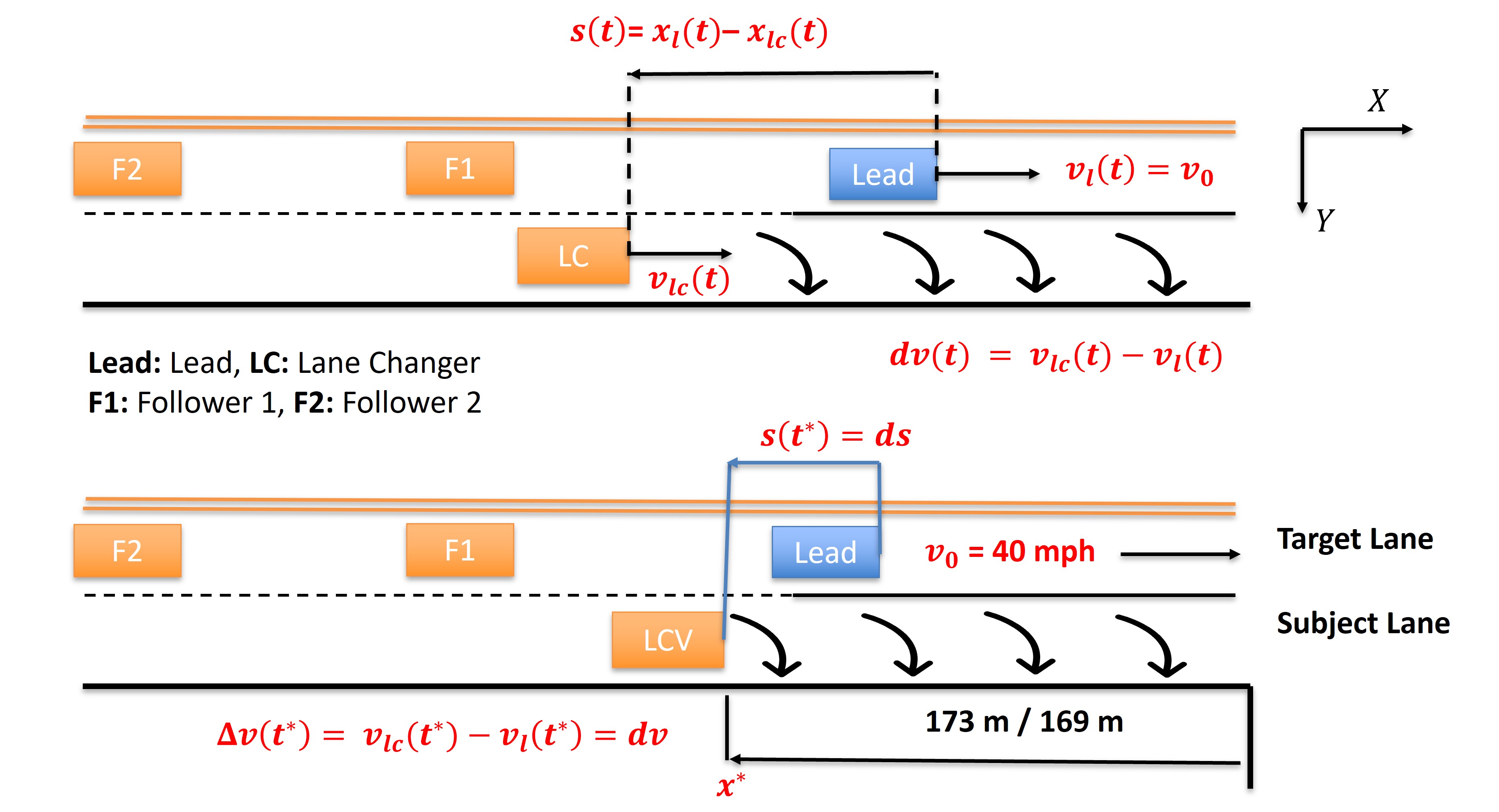}
    \caption{Vehicle configuration}
    \label{fig:Vehicle Configuration}
\end{figure}}

\newcommand{\ReferenceMaps}{
\begin{figure}[H]
    \centering
    \includegraphics[width=\linewidth]{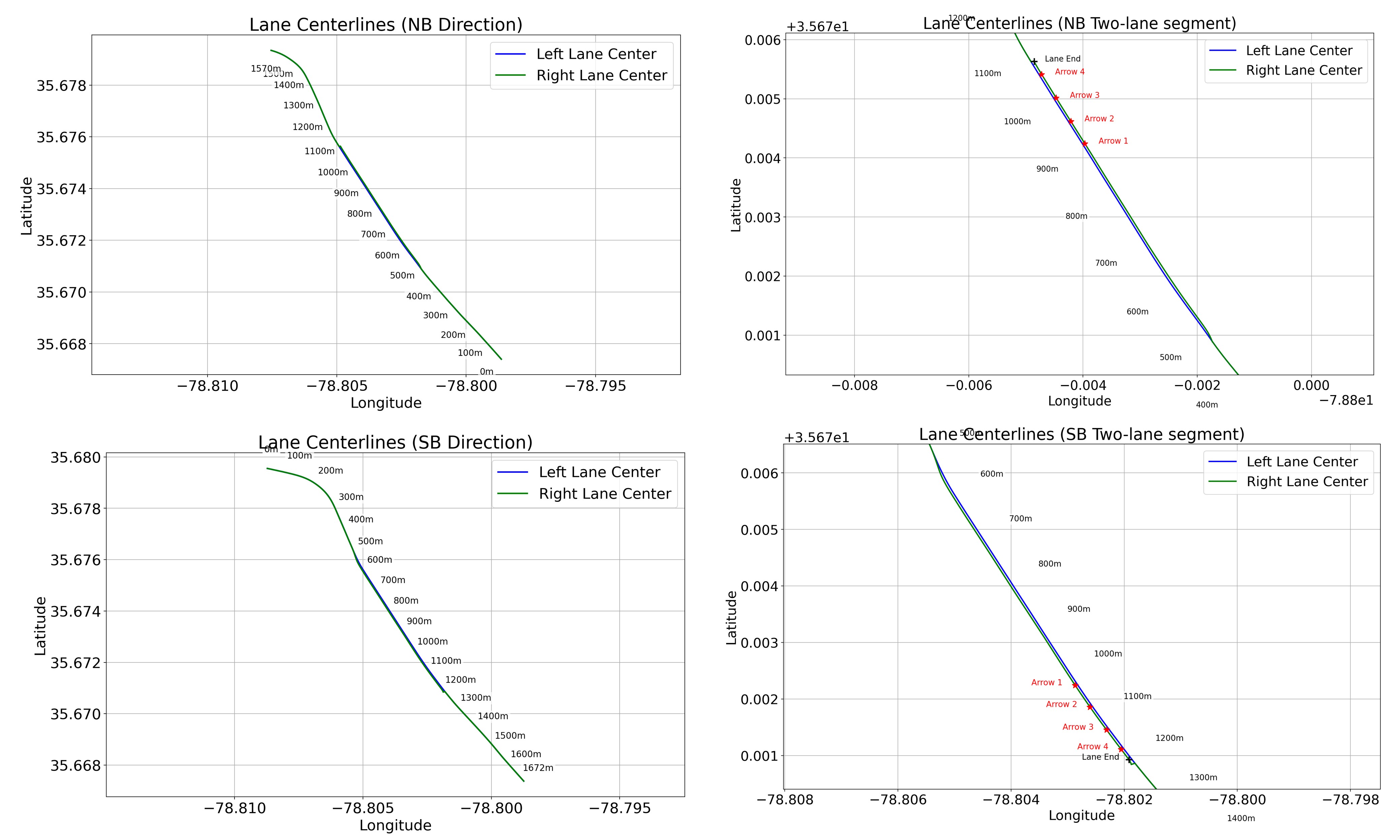}
    \caption{Lane Centerlines NB and SB direction with intermittent exclusive right-turn segment}
    \label{fig:Reference Maps}
\end{figure}}

\newcommand{\NGSIMtotalgap}{
\begin{figure}[H]
    \centering
    \includegraphics[width=\linewidth]{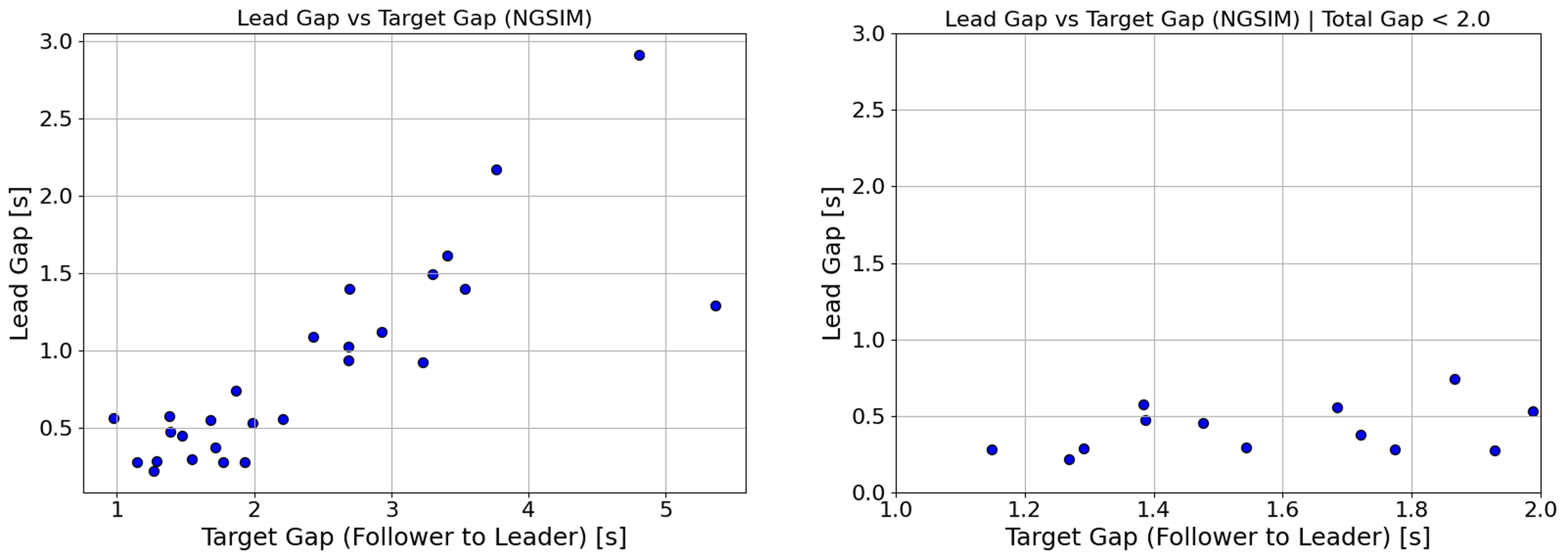}
    \caption{NGSIM Lead Gap vs Target Gap; (1) US101 Lane 6 to Lane 5 mandatory lane change cases (2) Only constrained target Gap (< 5 seconds) (3) Sample Size = 28}
    \label{fig:NGSIMtotalgap}
\end{figure}}

\newcommand{\IndependentVariablesSelection}{
\begin{figure}[H]
  \centering
  \includegraphics[width=\linewidth]{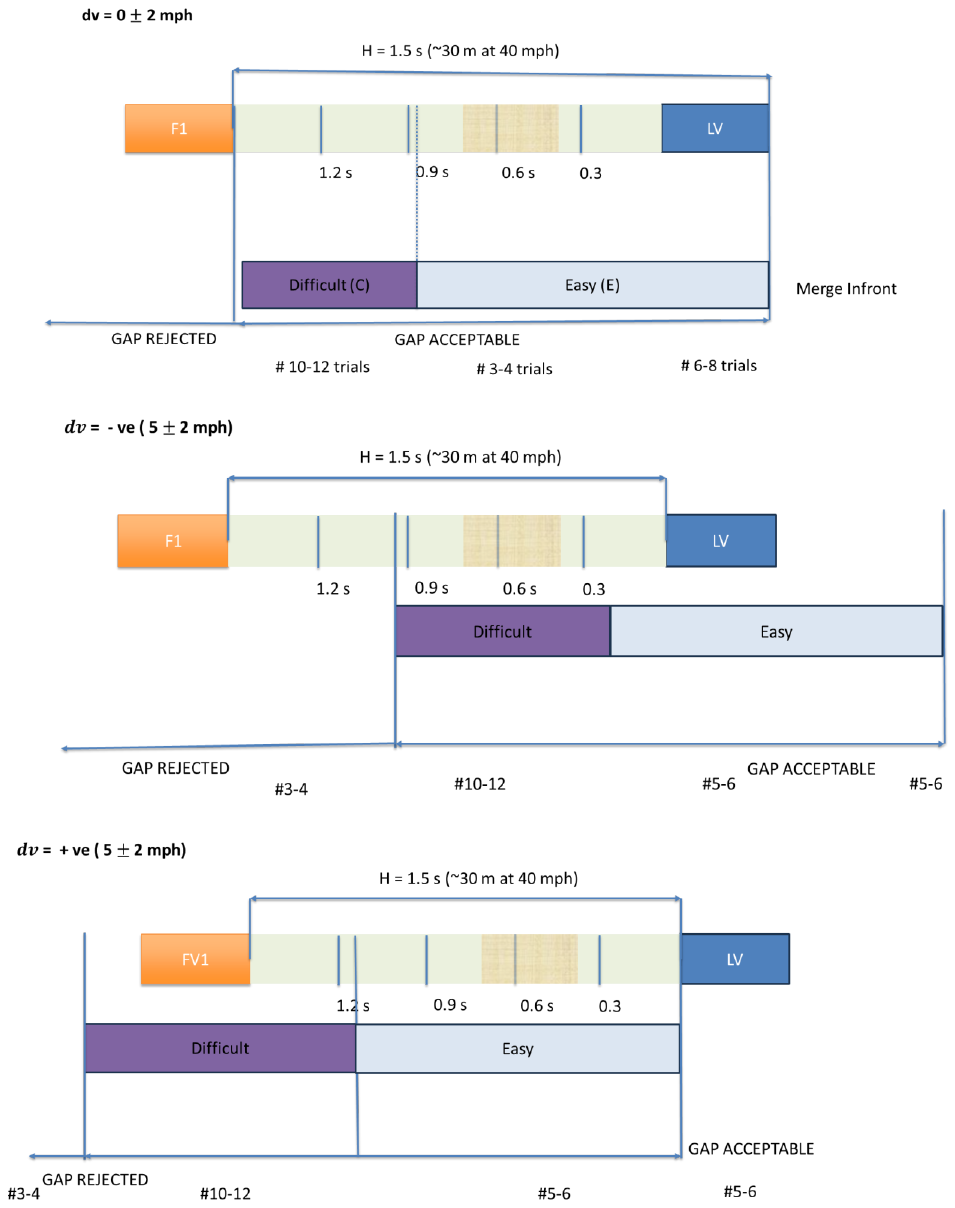}
  \caption{Selection of Independent Variables}
  \label{fig:independent_variables_selection}
\end{figure}}

\newcommand{\OverlayTrajectories}{
\begin{figure}[H]
    \centering
    \includegraphics[ height = 0.8 \paperheight]{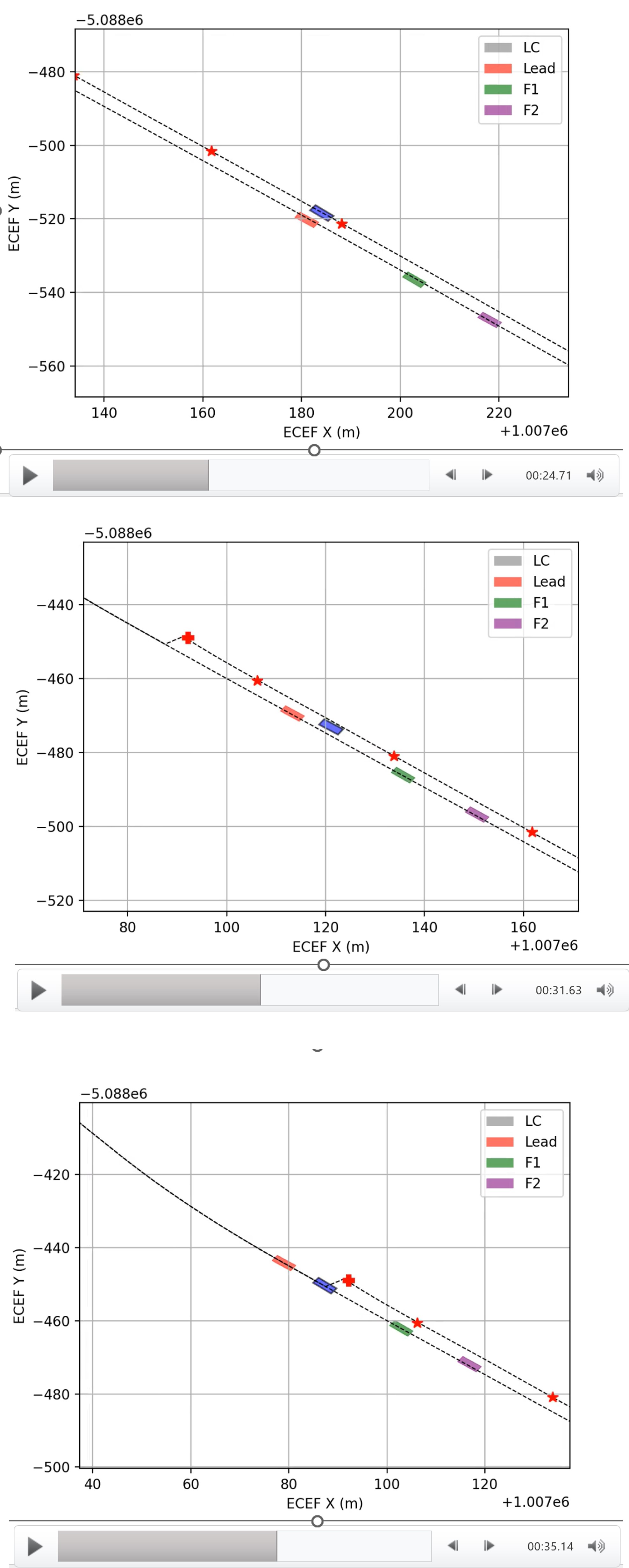}
    \caption{(Sample Example 1) Vehicles position and lane centerline at three timestamp t= 24 s, t = 31 s and t = 35 s. The red"asterisk" indicate the right-turn pavement markings and red "plus" sign indicates the end of exclusive right-turn lane in NB direction.}
    \label{fig:OverlayTrajectories}
\end{figure}}

\newcommand{\IndependentVariables}{
\begin{figure}[H]
    \centering
    \includegraphics[width=\linewidth]{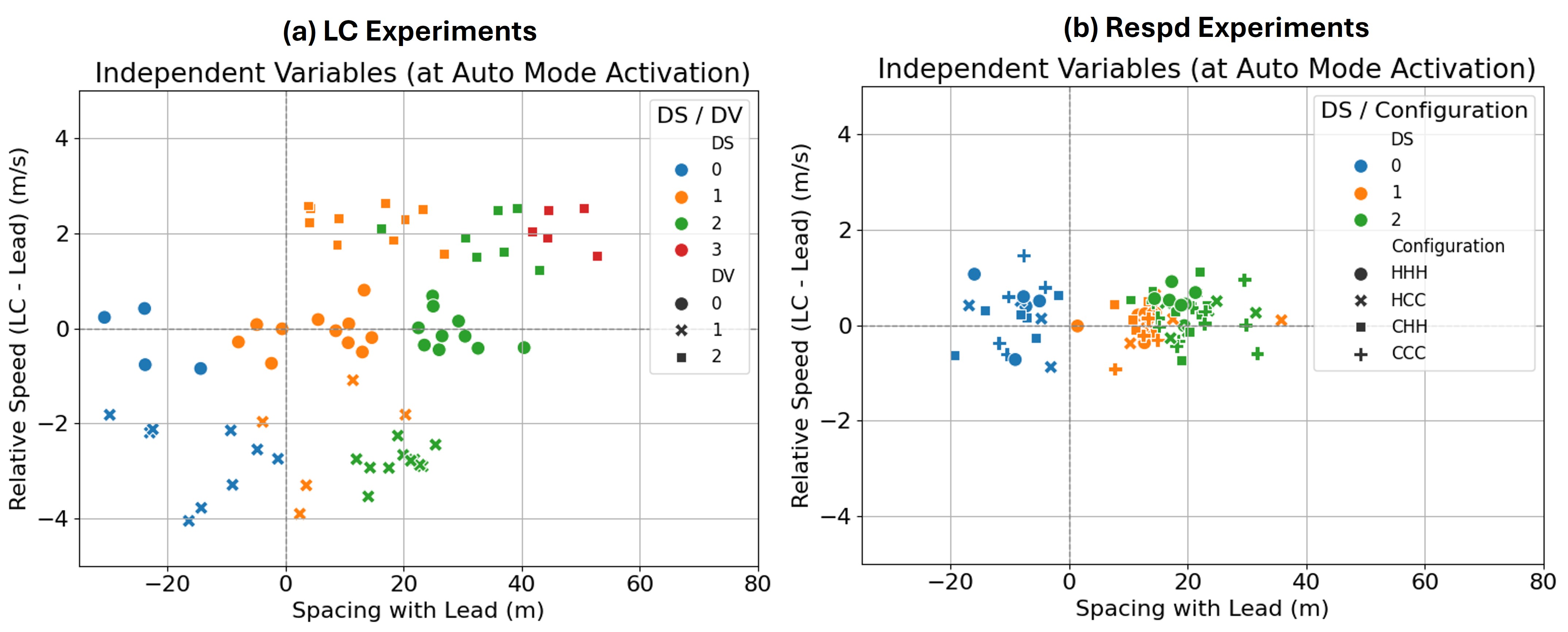}
    \caption{Independent Variables}
    \label{fig:IndependentVariables}
\end{figure}}

\newcommand{\SampleExampleLC}{
\begin{figure}[H]
    \centering
    \includegraphics[ height = 22cm]{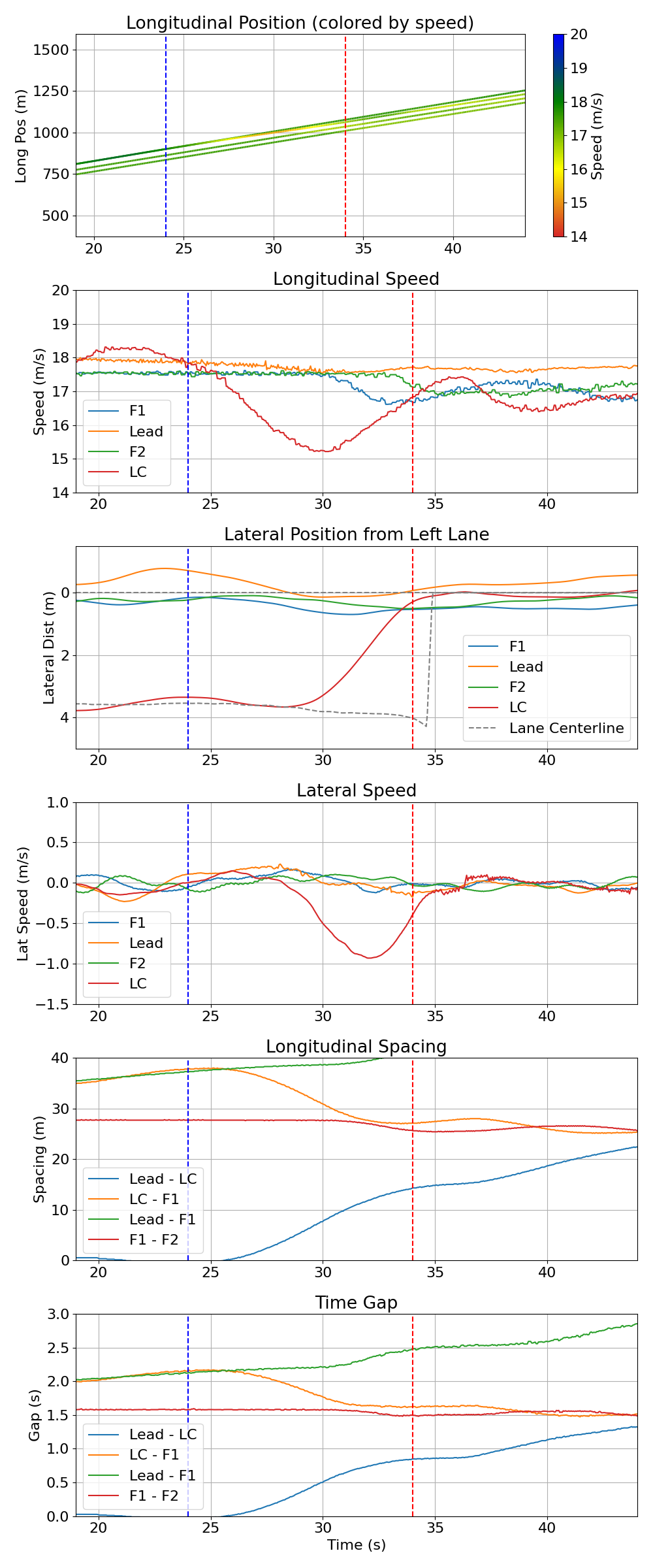}
    \caption{Sample Example 1 from LC Experiment (DS = 1, DV = 0)}
    \label{fig:Sample Example LC Experiments}
\end{figure}}

\newcommand{\OverlayTrajectoriesT}{
\begin{figure}[H]
    \centering
    \includegraphics[ height = 0.8 \paperheight]{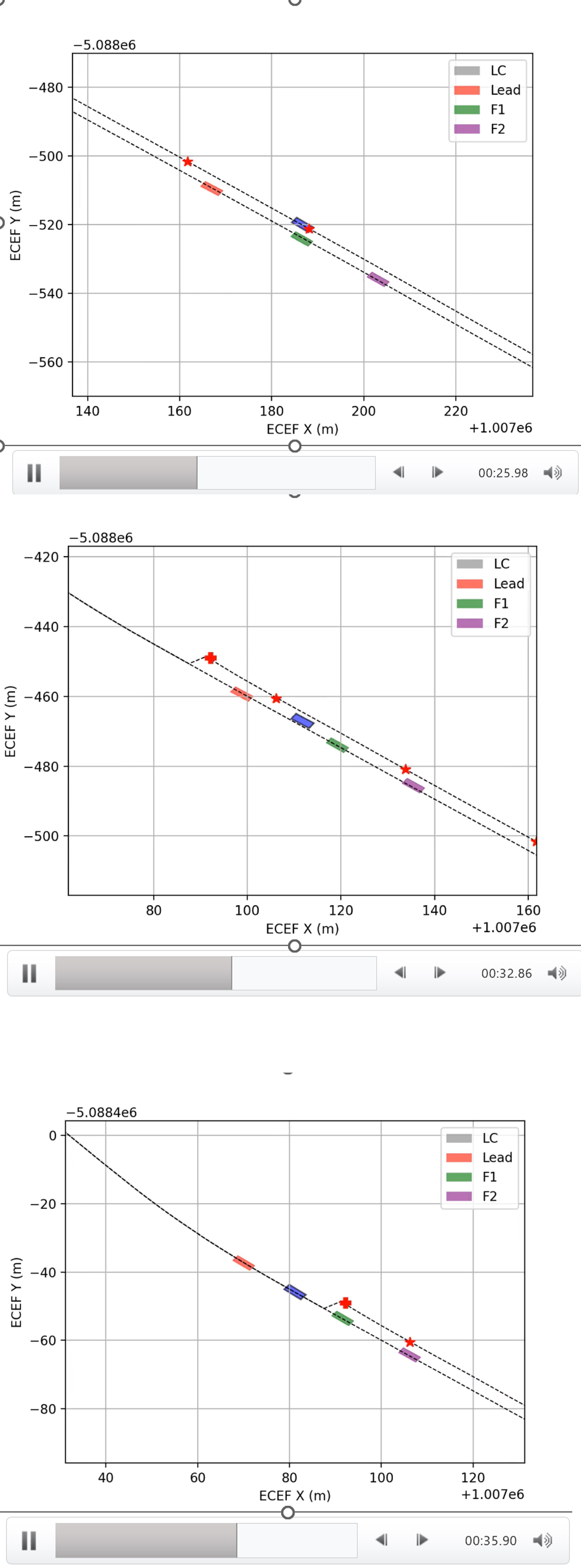}
    \caption{(Sample Example 2) Vehicles position and lane centerline at three timestamp t= 26 s, t = 33 s and t = 36 s. The red"asterisk" indicate the right-turn pavement markings and red "plus" sign indicates the end of exclusive right-turn lane in NB direction.}
    \label{fig:OverlayTrajectories 2}
\end{figure}}

\newcommand{\SampleExampleLCT}{
\begin{figure}[H]
    \centering
    \includegraphics[ height = 22cm]{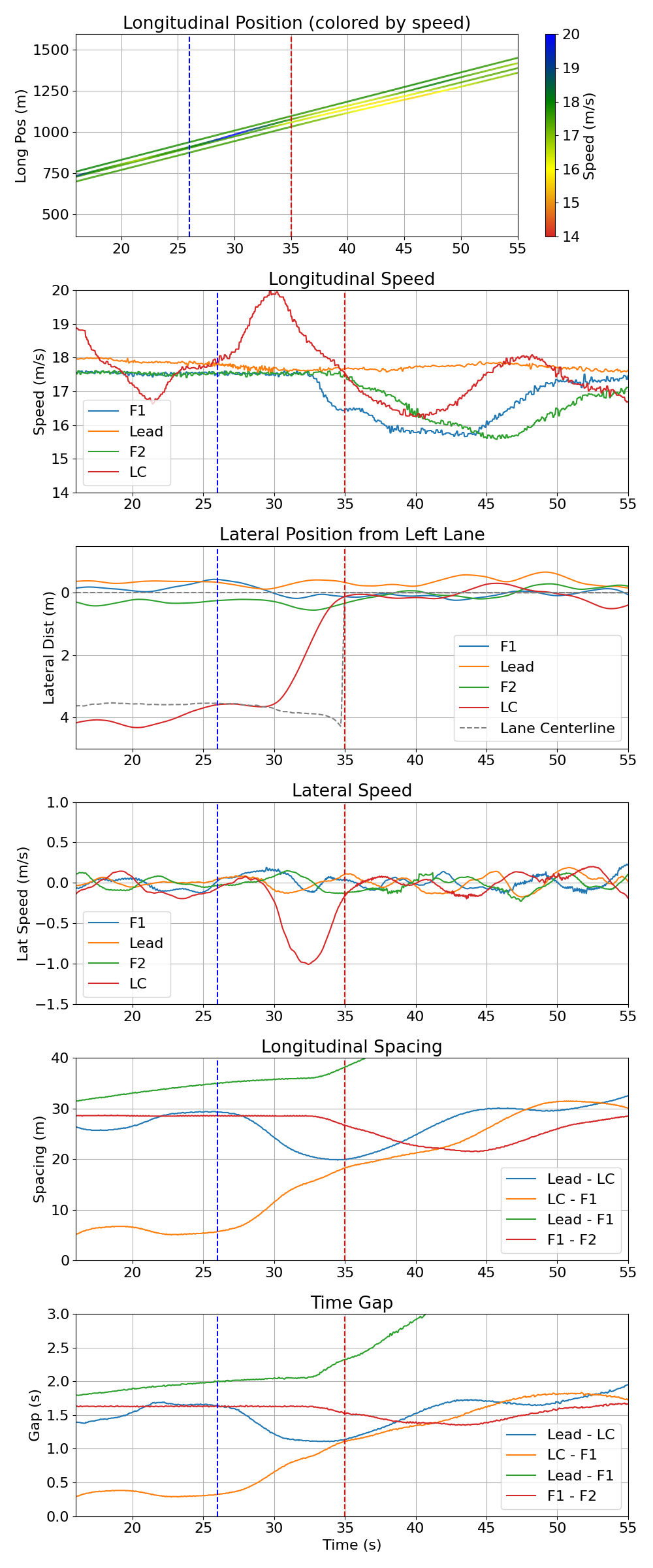}
    \caption{Sample Example 2 from LC Experiment (DS = 2, DV = 0)}
    \label{fig:Sample Example LC Experiments 2}
\end{figure}}

\newcommand{\SampleExampleRespd}{
\begin{figure}[H]
    \centering
    \includegraphics[ height = 22cm]{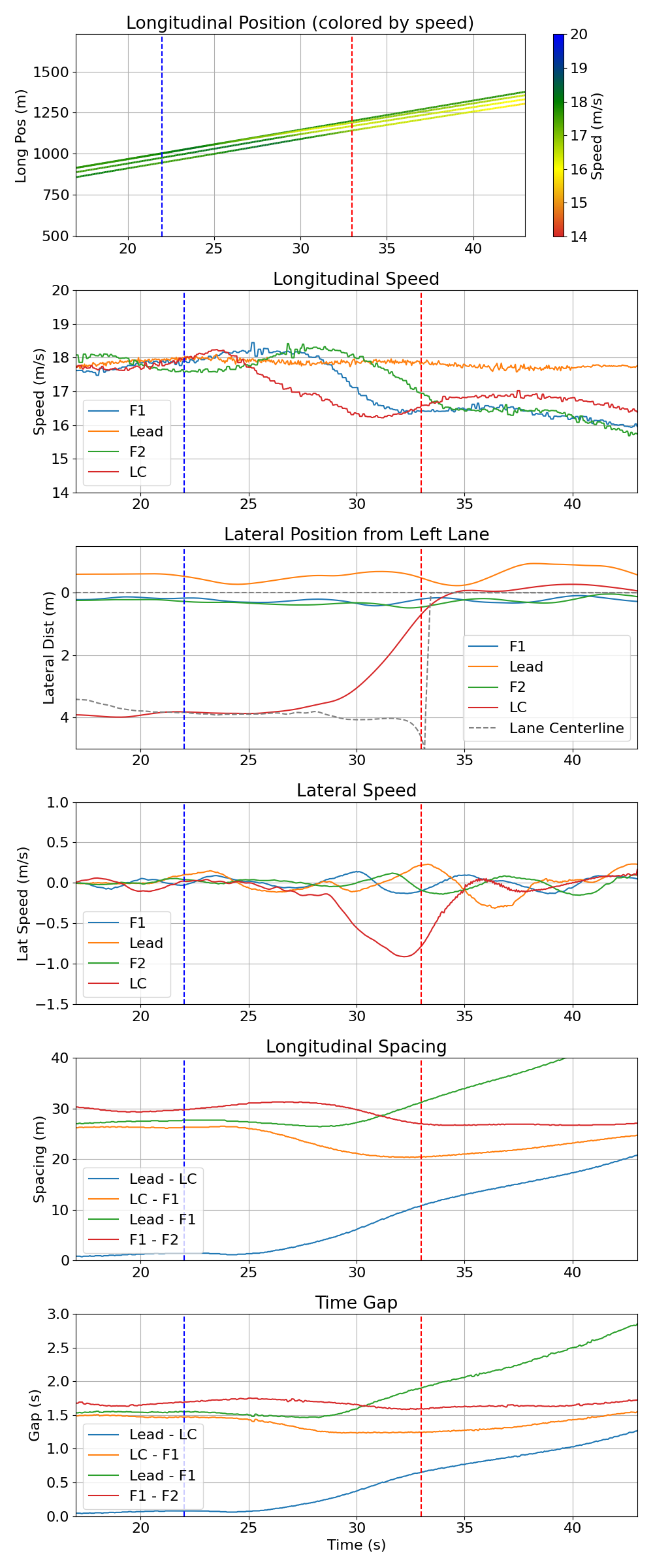}
    \caption{Sample Example from Respd Experiment (DS = 0, Config: HHH)}
    \label{fig:Sample Example Respd Experiments}
\end{figure}}

\newcommand{\TargetVehiclesSpeedLC}{
\begin{figure}[H]
    \centering
    \includegraphics[width=\linewidth]{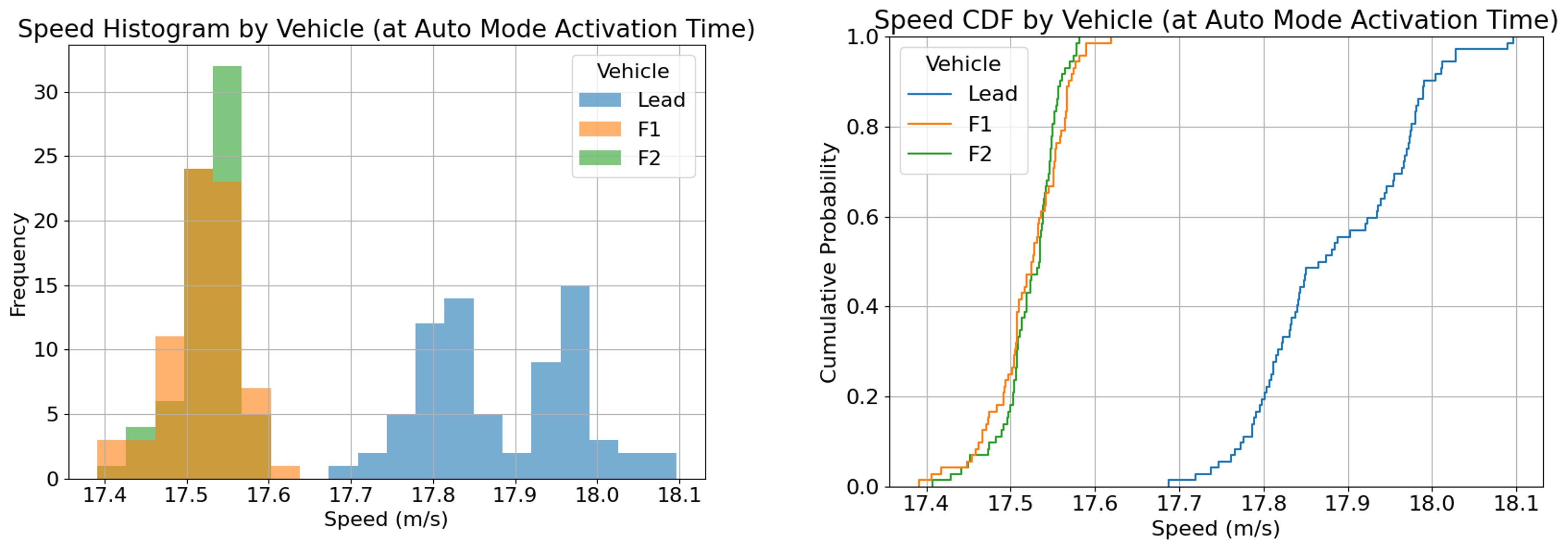}
    \caption{Target Vehicles Speed Distribution in LC Experiments (\textbf{Lead Vehicle}: Mean: 17.89 m/s Std Dev: 0.09 m/s Median: 17.87 m/s 25th \%: 17.81 m/s 75th \%: 17.97 m/s Min: 17.69 m/s Max: 18.10 m/s 
\textbf{F1 Vehicle}: Mean: 17.52 m/s Std Dev: 0.04 m/s Median: 17.53 m/s 25th \%: 17.50 m/s 75th \%: 17.55 m/s Min: 17.39 m/s Max: 17.62 m/s 
\textbf{F2 Vehicle}: Mean: 17.52 m/s Std Dev: 0.04 m/s Median: 17.53 m/s 25th \%: 17.51 m/s 75th \%: 17.55 m/s Min: 17.41 m/s Max: 17.58 m/s
)}
    \label{fig:Target Vehicles Speed in LC Experiments}
\end{figure}}

\newcommand{\TargetGapLC}{
\begin{figure}[H]
  \centering
  \includegraphics[width=\linewidth]{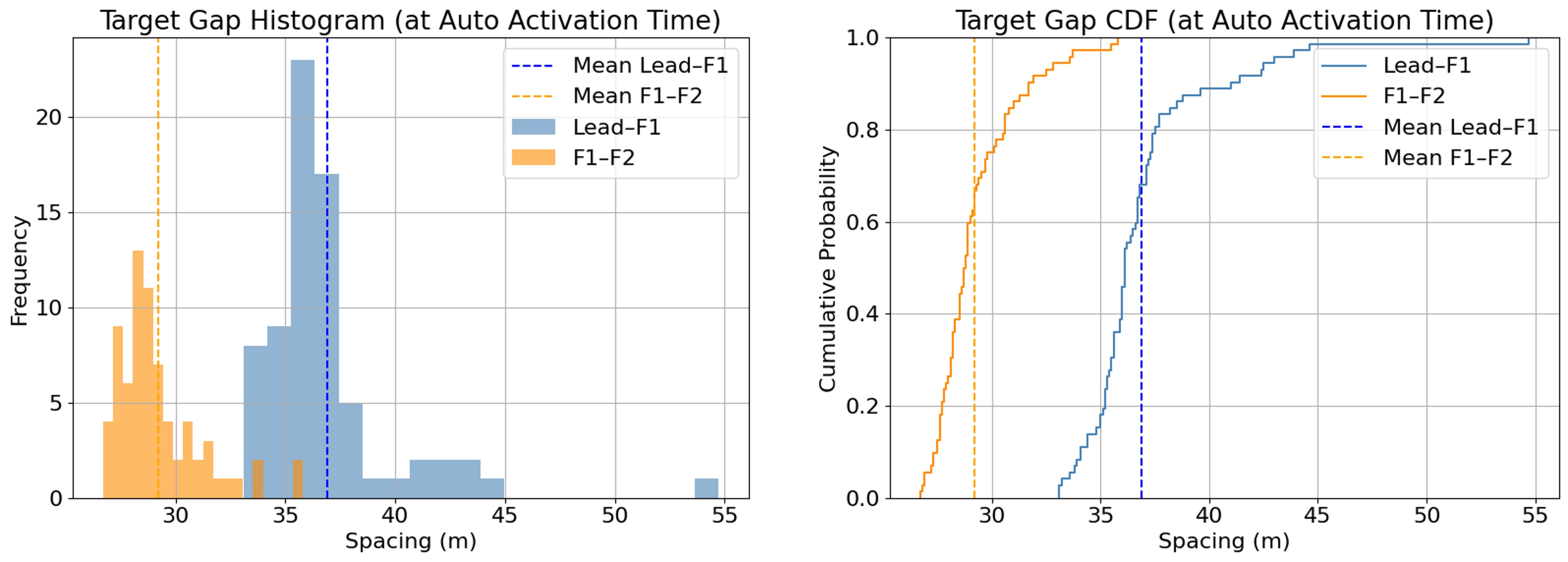}
  \caption{Target Gaps in LC Experiments measured at Auto Mode Activation Time (\textbf{Lead–F1 Spacing:} Mean: 36.90 m Std Dev: 3.21 m Median: 36.10 m 25th \%: 35.30 m 75th \%: 37.33 m \textbf{}\textbf{F1–F2 Spacing:} Mean: 29.22 m Std Dev: 1.92 m Median: 28.75 m 25th \%: 27.98 m 75th \%: 29.88 m
)}
  \label{fig:Target Gap in LC Experiment}
\end{figure}}

\newcommand{\TargetGapRespd}{
\begin{figure}[H]
  \centering
  \includegraphics[width=\linewidth]{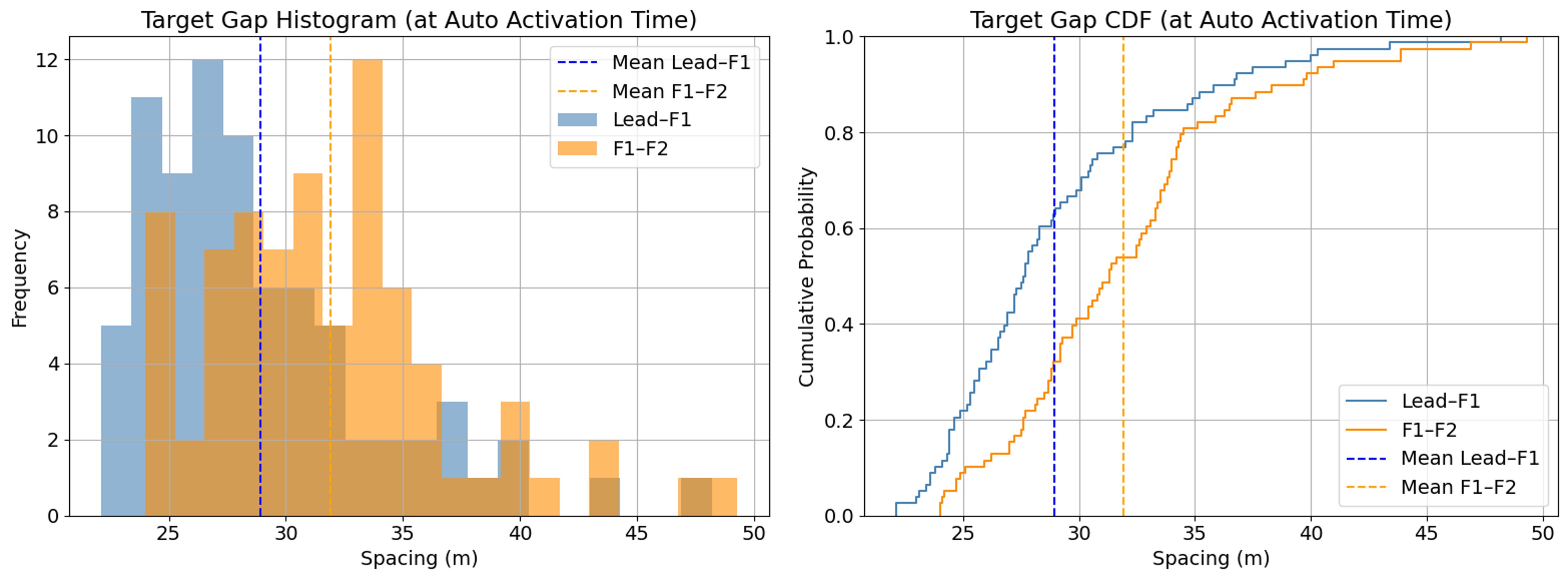}
  \caption{Target Gaps in Respd Experiments measured at Auto Mode Activation Time (\textbf{Lead–F1 Spacing}: Mean: 28.92 m Std Dev: 5.10 m Median: 27.65 m 25th \%: 25.35 m 75th \%: 30.75 m \textbf{F1–F2 Spacing}: Mean: 31.92 m Std Dev: 5.24 m Median: 31.30 m 25th \%: 28.55 m 75th \%: 34.15 m)
}
  \label{fig:Target Gap in Respd Experiment}
\end{figure}}

\newcommand{\TargetVehiclesSpeedRespd}{
\begin{figure}[H]
    \centering
    \includegraphics[width=\linewidth]{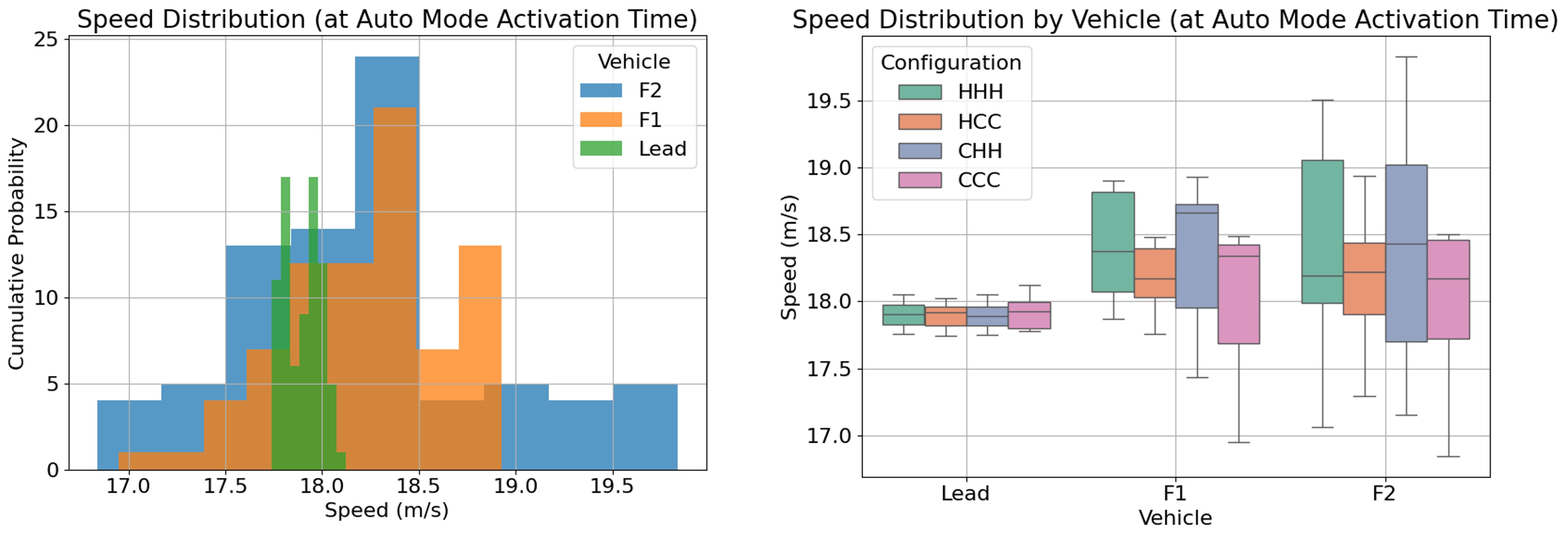}
    \caption{Target Vehicles Speed Distribution in Respd Experiments (\textbf{Lead Vehicle}: Mean: 17.90 m/s Std Dev: 0.09 m/s Median: 17.91 m/s 25th \%: 17.82 m/s 75th \%: 17.98 m/s Min: 17.74 m/s Max: 18.12 m/s \textbf{F1 Vehicle}: Mean: 18.24 m/s Std Dev: 0.43 m/s Median: 18.34 m/s 25th \%: 17.97 m/s 75th \%: 18.49 m/s Min: 16.95 m/s Max: 18.93 m/s F2 Vehicle: Mean: 18.24 m/s Std Dev: 0.67 m/s Median: 18.21 m/s 25th \%: 17.78 m/s 75th \%: 18.49 m/s Min: 16.84 m/s Max: 19.84 m/s
}
    \label{fig:Target Vehicles Speed in Respd Experiments}
\end{figure}}

\newcommand{\GapAcceptance}{
\begin{figure}[H]
    \centering
    \includegraphics[width=\linewidth]{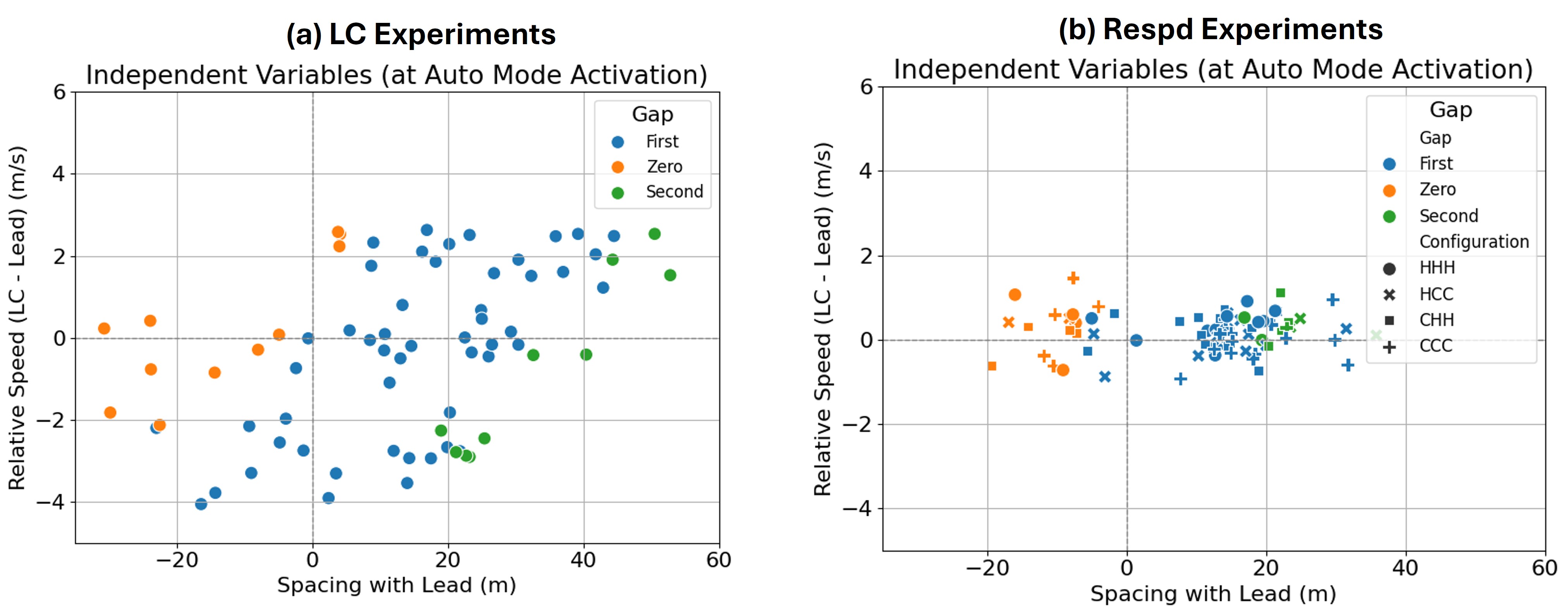}
    \caption{Accepted Gap across independent variables}
    \label{fig:Gap Acceptance}
\end{figure}}

\newcommand{\VehicleInstrumentation}{
\begin{table}[ht]
\footnotesize
\centering
\caption{Vehicle Specifications and Instrumentation Summary}
\label{tab:vehicle_instrumentation}
\renewcommand{\arraystretch}{1.2}
\resizebox{\textwidth}{!}{
\begin{tabular}{>{\raggedright\arraybackslash}p{2.3cm} >{\raggedright\arraybackslash}p{3.4cm} >{\raggedright\arraybackslash}p{2.7cm} >{\raggedright\arraybackslash}p{3.4cm} >{\raggedright\arraybackslash}p{3.2cm} >{\centering\arraybackslash}p{1.4cm}}
\toprule
\rowcolor{gray!20}
\textbf{Vehicle Type} & \textbf{System} & \textbf{Sensor / Device} & \textbf{Model} & \textbf{Accuracy (RMS)} & \textbf{Freq.} \\ 
\midrule

\multirow{2}{*}{Lead} 
  & \multirow{2}{=}{ACC} 
  & INS & NovAtel Dual-Antenna RTK-GPS & 
  \begin{tabular}[t]{@{}l@{}}
    Position: 0.01 m\\
    Heading: 0.09° @ 2 m \\
    Velocity: 0.02 m/s
  \end{tabular} & 20 Hz \\
  & & & & & \\

\midrule

\multirow{3}{*}{LC \& F1} 
  & \multirow{3}{=}{ tAV} 
  & INS  & Inertial Labs INS-DM (RTK) & 
  \begin{tabular}[t]{@{}l@{}}
    Position: 0.01 m\\
    Velocity: 0.01 m/s\\
    Heading: 0.05°
  \end{tabular} & 20 Hz \\
  & & Cameras & 4× Sony ZV-1F & 1920×1080, 84° FOV & 30 fps \\
  & & CAN Device & VLinker FS Bluetooth & --- & --- \\

\midrule

\multirow{3}{*}{F2} 
  & \multirow{3}{=}{tAV} 
  & INS & Inertial Labs INS-DU (RTK) & 
  \begin{tabular}[t]{@{}l@{}}
    Position: 0.01 m\\
    Velocity: 0.05 m/s\\
    Heading: 0.10°
  \end{tabular} & 20 Hz \\
  & & Cameras & 2× Sony ZV-1F & 1920×1080, 84° FOV & 30 fps \\
  & & CAN Device & VLinker FS Bluetooth & --- & --- \\

\bottomrule
\end{tabular}}
\end{table}
}
\newcommand{\ParameterSettings}{
\begin{table}[H]
\centering
\caption{Driving Settings (LC and Respd Experiments)}
\label{tab:lc_respd_settings}
\renewcommand{\arraystretch}{1.2}
\begin{tabular}{|p{4.1cm}|p{2.4cm}|p{2.4cm}|p{2.4cm}|p{2.4cm}|}
\hline
\rowcolor{red!20}
\textbf{Parameters} & \textbf{LC} & \textbf{LV} & \textbf{F1} & \textbf{F2} \\
\hline
\multicolumn{5}{|c|}{\cellcolor{gray!15}\textbf{LC Experiment}} \\
\hline
Driving Mode & Auto & ACC\textsuperscript{1} & ACC & ACC \\
Driving Style & Hurry & NA & Headway 2\textsuperscript{2} & Headway 2\textsuperscript{2} \\
Set Speed & Speed limit & 40 mph & 40 mph & 40 mph \\
Speed Offset & 40\%\textsuperscript{3} & NA & 0\% & 0\% \\
\hline
\multicolumn{5}{|c|}{\cellcolor{gray!15}\textbf{Respd Experiment}} \\
\hline
Driving Mode & Auto & ACC & Auto & Auto \\
Driving Style & Variable\textsuperscript{4} & NA & Variable & Variable \\
Set Speed & Speed limit & 40 mph & Speed limit & Speed limit \\
Speed Offset & Variable & NA &  & Variable \\
\hline
\end{tabular}
\vspace{1em}

\begin{flushleft}
\small
\textsuperscript{1} ACC: Lead Vehicle is not tAV. \\
\textsuperscript{2} Headway 2: Minimum Headway in ACC mode \\
\textsuperscript{3} Speed Offset (from detected speed limit): 20\% for Chill Mode and 40\% for Hurry Mode \\
\textsuperscript{4} Variable: Either \textit{Hurry} or \textit{Chill} mode per trial. (see configurations HHH/HCC/CHH/CCC) \\
\end{flushleft}
\end{table}
}

\newcommand{\SampleSize}{
\begin{table}[H]
\centering
\caption{Sample Size Distribution for LC and Respd Experiments}
\label{tab:sample_distribution_combined}
\renewcommand{\arraystretch}{1.2}
\begin{tabular}{|c|c|c|}
\hline
\rowcolor{red!20}
\multicolumn{3}{|c|}{\textbf{LC Experiment — Sample Size by Discrete Spacing (DS) and Discrete Velocity (DV)}} \\
\hline
\rowcolor{red!10}
\textbf{DS} & \textbf{DV} & \textbf{Sample Size (72)} \\
\hline
0 & 0 & 4 \\
1 & 0 & 11 \\
2 & 0 & 10 \\
0 & 1 & 9 \\
1 & 1 & 5 \\
2 & 1 & 12 \\
1 & 2 & 10 \\
2 & 2 & 6 \\
3 & 2 & 5 \\
\hline
\multicolumn{3}{|c|}{\cellcolor{gray!20}\textbf{Respd Experiment — Sample Size by Configuration and Discrete Spacing (DS)}} \\
\hline
\rowcolor{red!10}
\textbf{Configuration} & \textbf{DS} & \textbf{Sample Size (80)} \\
\hline
HHH & 0 & 5 \\
HHH & 1 & 6 \\
HHH & 2 & 8 \\
HCC & 0 & 4 \\
HCC & 1 & 7 \\
HCC & 2 & 9 \\
CHH & 0 & 6 \\
CHH & 1 & 5 \\
CHH & 2 & 8 \\
CCC & 0 & 6 \\
CCC & 1 & 5 \\
CCC & 2 & 11 \\
\hline
\end{tabular}
\end{table}
}
\section{Introduction}

    
Lane-changing (LC) maneuvers are frequent yet safety-critical components of driving, involving tactical decisions such as gap selection and precise execution through speed modulation and spatial coordination in dynamic traffic environments. The adverse effects of LC on both traffic capacity and road safety have been well-documented in prior research \cite{Zheng2014RecentChanging,Shi2021EmpiricalSettings}. 

With ongoing advancements in vehicle automation, transitional autonomous vehicles (tAVs), which operate beyond conventional SAE Level 1–2 automation but are not fully autonomous—are increasingly present on roadways \cite{Tesla2025TeslaAutopilot, SuperCruise2025SuperCruise}. For instance, approximately 5 million Tesla vehicles have been sold globally, of which about 500,000 are equipped with supervised self-driving capabilities \cite{Buyacar2025Buyacar, Apnews2025Apnews}. These systems enable tAVs to execute automated lane-changing maneuvers. When operating alongside human-driven vehicles (HDVs), tAVs may give rise to novel interaction patterns, with potential implications for traffic safety and flow stability.

Despite the growing presence of tAVs on public roads, empirical evidence on their performance during LC maneuvers in mixed traffic remains limited. Some prior studies have explored  LC behavior of autonomous vehicles (AVs) using real-world datasets. For instance,  \cite{Ali2024InvestigatingDataset} analyzed the Waymo Open Dataset to extract discretionary LC trajectories of AVs (147 samples) and compare them with those of human-driven vehicles (HDVs). Their findings suggest that AVs tend to exhibit longer lane-change durations, larger lead and lag gaps, and lower acceleration variability—indicators of more conservative and potentially safer LC behavior. In a follow-up study, \cite{Ali2025AutonomousFollower} investigated how AV lane changes affect the behavior of following human drivers. Using the surrogate safety measure (SSM) known as the deceleration rate to avoid a crash (DRAC), the study observed a heightened risk of rear-end collisions when the lane-changing vehicle was an AV compared to an HDV. These studies highlight the need for deeper insights into AV–HDV interactions during LCs, particularly in the transitional phase of automation. However, there remains a lack of high-resolution, structured datasets focused specifically on tAV behavior during both LC execution and response scenarios—especially under controlled experimental conditions.

Ammourah et al. \cite{Ammourah2024IntroductionExtraction} made notable progress in the tAV dataset domain by introducing the Third Generation Simulation Dataset (TGSIM), which aims to capture tAV–human interactions in naturalistic driving contexts. TGSIM encompasses data from Level 1, 2, and 3 automated vehicles operating in diverse highway environments, including mandatory lane-changing scenarios and urban intersections. The data were collected using three methods: (1) fixed-location aerial videography via helicopter hovering over a target segment, (2) moving aerial videography where a helicopter follows the vehicle across longer distances, and (3) infrastructure-based videography using multiple overlapping cameras mounted on overpasses. The dataset is designed to support analyses of traffic flow, string stability, and lane-changing behavior. However, detailed trajectory-based investigations are still underway.

 In this paper, we present controlled empirical experiments with tAVs focused on mandatory lane-changing scenarios. This dataset enables a deeper understanding of the interaction dynamics of tAVs both as lane-change initiators and as responders. The experiments were conducted at an exclusive right-turn lane, providing mandatory lane-change scenario. The trajectory dataset was collected using high-precision RTK-GPS devices. The dataset, referred to as North Carolina Transitional Autonomous Vehicle Lane-Changing (NC-tALC) dataset,  consists of two experimental setups: (1) LC Experiments, where a tAV executes lane changes in the presence of adaptive cruise control (ACC)-equipped target vehicles, providing insights into tAV LC behavior; and (2) Responding (Respd) Experiments, where two tAVs act as following vehicles responding to cut-in maneuvers performed by another tAV. Notably, the same tAVs operated in ACC Mode during the LC experiments and in Autonomous (Auto) Mode during the Respd experiments. 

The main contribution of this paper is the NC‑tALC dataset: 152 controlled scenarios - 72 LC cases and 80 Respd cases - collected under systematically varied LC decision‑making factors. In LC experiments, we vary relative speed and relative spacing; in Respd experiments, we vary follower driving style (aggressive vs. conservative) and LC vehicle relative spacing. This targeted variation is difficult to obtain from naturalistic datasets. The experimental framework is reproducible and scalable, supporting expansion to additional variables, vehicles, and locations. Unlike large naturalistic datasets (e.g., Waymo, TGSIM), NC‑tALC provides repeatable mandatory LCs and paired follower cut‑in responses trajectories enabling controlled cause–effect analyses seldom possible in naturalistic datasets.

The NC-tALC dataset has the potential to inform the development of safer and more efficient AV lane-changing algorithms, improve interaction strategies in mixed-traffic environments, and support the design of robust autonomous driving systems.


\section{Methods and Materials}

    

This section outlines the logistical framework of the empirical study. It begins by describing the experiment location and road geometry, followed by an overview of the vehicle setup and instrumentation. It then briefly introduces the pre-LC on-field execution coordination strategy to achieve the experimental scenario, and finally covers the reference map of test site.

\subsection{Experiment Location}



The experiments were conducted on a 600-meter stretch of Sunset Lake Road in Apex, North Carolina, USA. Sunset Lake Road is a single-lane major collector where two lanes are intermittently provided to accommodate exclusive right-turn movements (see Figures ~\ref{fig:Test Site} and \ref{fig:Site Layout}). The selected segment features two lanes in both the northbound (NB) and southbound (SB) directions, with the rightmost lane in each direction designated exclusively for right turns. The right-turn pavement markings extend approximately 173 meters in the NB direction and 169 meters in the SB direction, offering comparable geometric layouts in both directions. Both NB and SB lanes were utilized for testing. The posted speed limit on the segment is 45 mph. Experiments were conducted over a 10-day period during daylight hours (10:00 AM to 5:00 PM), under clear and sunny weather conditions to ensure optimal visibility and environmental consistency. Importantly, the site allowed for real-world testing in live traffic without requiring lane closures or causing disruptions to other road users. The two-lane segment is nearly straight (Average heading: \(331.45\degree\), Min: \(328\degree\), Max:\(334\degree\)) with negligible grade (\(< 2\%\)) ; lanes are 3.7 m with no shoulder width; no signals present within the study segment. 

\figTestSite
\SiteLayout

\subsection{Vehicles and Instrumentation}


The experimental setup involved a four-vehicle configuration consisting of one LC vehicle  and a three-vehicle platoon in the target lane, comprising a lead vehicle, a first follower (F1), and a second follower (F2), as illustrated in Figure~\ref{fig:Vehicle Configuration}. The Lead vehicle was equipped with ACC, while the LC, F1, and F2 vehicles were tAVs exhibiting autonomous driving capabilities. Each vehicle was instrumented with a high-precision Inertial Navigation System (INS) that integrated Real-Time Kinematic (RTK) positioning with a Global Navigation Satellite System (GNSS), accelerometer, and gyroscope. The INS provided time-synchronized outputs of position, speed, heading, and acceleration at a sampling rate of 20 Hz. As per manufacturer specification, the INS device achieves centimeter-level positional accuracy and a speed accuracy of 0.01 m/s. In addition, each tAV was equipped with four in-cabin cameras to record both environmental and dashboard data (see Table~\ref{tab:vehicle_instrumentation} for details).

\figVehicleConfig
\VehicleInstrumentation

\subsection{Execution Strategy}

Prior to entering the two-lane segment, all four vehicles— LC, Lead, F1, and F2—were cruising in a single-lane formation at a steady speed of 35 mph. The LC was manually driven at this speed, while the Lead, F1, and F2 vehicles operated in ACC or Auto Mode, with a set speed of 40 mph (but cruising at 35 mph). F1 and F2 shared identical ACC/Auto settings, which were distinct from those of the Lead vehicle (detailed in the following section).

Upon entering the two-lane stretch, the LC driver manually merged into the exclusive right-turn lane. Once the LC vehicle transitioned, the Lead vehicle accelerated from 35 mph to its set speed of 40 mph, followed by F1 and F2. The LC vehicle driver then modulated its speed and spacing relative to the Lead vehicle to match predefined experimental conditions for relative speed and spacing . At the beginning of the right-turn pavement markings, the LC driver activated Auto Mode, and tAV took over immediately. On recognizing the exclusive right-turn lane, tAV perform LC maneuver to left lane. To ensure consistent baseline conditions, the experiment was designed such that the Lead vehicle maintained a constant speed of 40 mph for at least 10–15 seconds before reaching the right-turn lane marking.

\subsection{Reference Maps}

A high-resolution reference map was developed for the experimental site to accurately represent the centerlines of the left and right lanes (see Figure \ref{fig:Reference Maps}). To construct this map, a transitional autonomous vehicle (tAV) was autonomously driven along each lane in both the northbound (NB) and southbound (SB) directions across three separate trials. The start and end times for each run were recorded, and data from the INS and synchronized video logs were collected. The INS data were originally recorded in a geographic coordinate system (GCS) and subsequently transformed into a local reference frame.

In the local frame, the centerline of the left lane was defined as the longitudinal reference axis. To generate coordinates in the local frame, geodesic distances between consecutive GPS points were computed using Python’s \texttt{geopy} library, which uses WGS-84 method to ensure spatial accuracy. Raw GPS trajectories from each lane were densified to a resolution of 0.05 meters using linear interpolation to preserve geometric detail. From this dense trajectory, evenly spaced reference points were sampled at 0.1-meter intervals to form smooth centerlines for both the left and right lanes. 

To align the two lanes spatially, the point on the left lane closest to the starting coordinate of the right lane was identified using geodesic distance. This alignment ensured synchronized coordinate systems and enabled direct point-to-point correspondence between the two lanes. At each aligned coordinate pair, the lateral spacing between lanes was computed, allowing for capture of local variations in road width at every 0.1-meter longitudinal step.

For each experimental trial, the trajectories of all vehicles—Lead, F1, F2, and the LC vehicle—were projected onto the left lane centerline, which served as the common longitudinal reference axis. At each timestamp, the nearest reference coordinate was identified, and the vehicle’s longitudinal and lateral positions were calculated relative to this point. Longitudinal position indicated the vehicle’s location along the lane, while lateral position represented its offset from the centerline.

To reduce GPS noise and referencing artifacts, a 20-point moving average (equivalent to one second) was applied to both longitudinal and lateral position time series. Vehicle speeds were computed by differentiating the smoothed longitudinal positions and validated against speed profiles from the INS to ensure consistency. This unified spatial referencing framework allowed for consistent spatial and temporal comparisons across all vehicles and trials, enabling detailed behavior analysis.

\ReferenceMaps

\section{NC-tALC Experiments}

This section describes the experimental design and divided into two subsections: (a) Lane-Changing (LC) Experiments, and (b) Responding (Respd) Experiments. For each experiment set, we outline the driving settings, independent variables that were systematically varied, and the rationale behind their selection. The section concludes with a summary of the sample distribution across different parameter combinations in both experiment types.

\subsection{LC Experiments}

The LC experiments are specifically designed to examine the tAV LC decision and execution behavior under varying initial conditions. These scenarios reflect a range of real-world interactions that occur during lane change execution, enabling detailed investigation into how different relative spacing and speed conditions influence tAV behavior.

The experimental factors are broadly categorized into two types: Driving settings (see LC experiments driving settings in Table~\ref{tab:lc_respd_settings})  and independent variables. The driving settings include driving Mode, driving style, set speed, and speed offset. 

\subsubsection{Driving Settings}

Driving Mode refers to whether the vehicle operates in ACC Mode or Auto Mode. Driving style applies to vehicles in Auto Mode and represents different levels of driving urgency, as defined by the tAV system: \textit{Chill}, \textit{Standard}, and \textit{Hurry}.\textit{ Chill} provides a more relaxed driving style with minimal lane changes and \textit{Hurry} drives with more urgency. We believe that this embedded urgency in the driving profile may influence the tAV's lane-changing behavior and trajectory characteristics under Auto Mode. Specifically, \textit{Hurry }Mode potentially make tAV aggressive, whereas \textit{Chill} Mode keeps it conservative. In the LC experiments, we used the \textit{Hurry} setting for LC vehicle to represent an aggressive driving configuration. This was important to observe how the tAV behaves when it is programmed to change lanes more actively. 

Set speed refers to the maximum cruising speed that a vehicle is allowed to reach under equilibrium conditions. Speed offset is an additional setting that allows the vehicle to exceed the detected speed limit by a certain percentage, if needed, to match the flow of traffic. For example, a positive speed offset enables the tAV to temporarily drive faster than the detected speed limit when required, such as during a lane change. 

In the LC experiments, F1 and F2 vehicles had a speed offset of 0\%, meaning they strictly followed ACC driving maximum speed limit.  In contrast, the LC vehicle was assigned a set speed equal to the speed limit, but with a 40\% speed offset. This higher offset allowed the LC tAV to accelerate beyond the limit if necessary, providing the flexibility to complete mandatory lane changes more effectively in real traffic conditions.

\subsubsection{Independent Variables: Relative Spacing and Speed}

The independent variables in the LC experiments are \textbf{relative spacing} \((s(t) = ds)\) and \textbf{relative speed} \((\Delta v(t) = dv)\), measured at the location \(x^*\) and time \(t^*\) when Auto Mode is activated (see Figure~\ref{fig:Vehicle Configuration}). For simplicity, we refer to these conditions jointly as \textit{ds-dv} throughout the remainder of the paper. To isolate the effects of \textit{ds} and \textit{dv}, the Auto Mode activation point \((x^*)\) was fixed across trials. This approach avoids introducing a third variable, such as the remaining distance available to complete the lane change. Further, it avoids any potential delay in activating tAV LC module. 

Previous lane-changing studies have shown that relative spacing and relative speed with respect to surrounding vehicles—particularly the target leader and follower—are critical factors influencing lane change decisions and execution profiles~\cite{Zheng2014RecentChanging,Ali2018ConnectivitysStudy}. Therefore, our experimental design systematically varies these two parameters to examine their effect on tAV behavior.

We selectively sampled combinations of \(ds-dv\) in a non-uniform manner, focusing on scenarios likely to challenge AV performance. The experimental design categorized these scenarios into easy and difficult zones, defined by conditions where human drivers typically find lane changes straightforward or hesitant, ensuring alignment with real-world driving contexts.

The selection of variables, as illustrated in Figure~\ref{fig:independent_variables_selection}, is based on a target gap of 1.5 seconds, divided into intervals of approximately 5 meters (one car length), equivalent to 0.3 seconds at 40 mph. We hypothesize that when relative speed is zero (dv = 0), larger lead gaps (ds > 0.9 s) may be more challenging for human drivers to accept than smaller gaps. This 0.9-second threshold is derived from mandatory LC scenario as per Next Generation Simulation (NGSIM) dataset \cite{FHWA2006NGSIM}, which suggest that human drivers typically accept lead gaps lower than 0.6 seconds when target gap (measured using spacing between the target leader and target follower divided by speed of target follower) is in range of 1-2 seconds (see Figure~\ref{fig:NGSIMtotalgap}). Adding a 0.3-second buffer (one car length) establishes the 0.9-second threshold. NGSIM data further suggests that drivers maintain larger gaps behind than in front during lane changes, indicating greater difficulty in assessing gaps near the following vehicle compared to the leading vehicle. Note that the above reasoning is based on a limited sample size, and the threshold values may vary. This analysis is presented here for demonstrative purposes to illustrate the existence of difficult and easy zones in tAV gap acceptance.

We further hypothesize that the challenging zone is influenced by dv. When dv > 0 (the LC vehicle is faster than the lead), the favorable speed difference reduces perceived risk, making larger ds values feel less difficult and expanding the easy zone toward higher gaps. Conversely, when dv < 0 (the LC vehicle is slower), merging near a F1 vehicle increases perceived risk, making smaller ds values more challenging. Consequently, the challenging zone shifts toward lower ds values, reflecting the interplay between speed and spacing in lane-change decisions.

\IndependentVariablesSelection
\NGSIMtotalgap

\textbf{Note:} As mentioned, the definitions of ``difficult'' and ``easy'' merging scenarios are based entirely on human driver behavior observed in small sample size. tAV, however, may apply different decision rules, risk tolerances, and sensor-based assessments that lead to different judgments about gap acceptability and maneuver feasibility.

From a field implementation perspective, ds and dv are discretized into the Discrete Spacing (DS) and Discrete Velocity (DV) categories, defined at Auto Mode activation (t*) for scenario execution and sampling.

\begin{enumerate}
    \item \textbf{Discrete Spacing Indicator (DS):} A discrete spacing variable representing the longitudinal position of the LC vehicle  with respect to the Lead vehicle, i.e. ds. Four spacing categories were defined:

    \begin{itemize}
        \item DS = 0: Ahead of Lead (ds < 0)
        \item DS = 1: Parallel to Lead and up to half target gap ( 0 < ds and ds < 20 m)
        \item DS = 2: From half target gap to F1 (ds >20 m and ds < 35 m)
        \item DS = 3: Behind F1 (ds > 35 m)
    \end{itemize}

    \item \textbf{Discrete Velocity Indicator (DV):} A discrete velocity variable representing the speed differential between the LC and Lead, i.e., dv. Three DV conditions were tested:

    \begin{itemize}
        \item DV = 0: Speed matched ($\Delta v = 0 \pm 2$ mph)
        \item DV = 1: LC vehicle slower ($\Delta v = -5 \pm 2$ mph)
        \item DV = 2: LC faster ($\Delta v = +5 \pm 2$ mph)
    \end{itemize}

\end{enumerate}

\ParameterSettings 

\subsection{Responding (Respd) Experiments}

The Respd experiments were designed to investigate the behavioral responses of target follower tAVs operating in Auto Mode when encountering a LC vehicle under varying spacing conditions (\(DV = 0\)).

Four combinations of AV driving styles were tested:

\begin{itemize}
    \item \textbf{HHH:} LC, F1, and F2 all in \textit{Hurry} Mode
    \item \textbf{HCC:} LC in \textit{Hurry}, F1 and F2 in \textit{Chill} Mode
    \item \textbf{CHH:} LC in \textit{Chill}, F1 and F2 in \textit{Hurry} Mode
    \item \textbf{CCC:} LC, F1, and F2 all in \textit{Chill} Mode
\end{itemize}

In \textit{Hurry} Mode, the speed offset was set to 40\%, while in \textit{Chill} Mode, it was set to 20\% from detected speed limit. The spacing conditions tested in the Respd experiments followed a similar sample distribution to that used in the LC experiments with \(DV = 0, |dv|< 2\ mph\).

These combinations were designed to evaluate how varying levels of driving assertiveness influence the target follower vehicle's response to a cut-in by the LC vehicle. In particular, the experiments aim to assess whether more aggressive driving styles (e.g., HHH) result in tighter following behavior or delayed deceleration, compared to more conservative settings (e.g., HCC or CCC). The LC vehicle always initiated the cut-in under consistent spacing and speed conditions, enabling isolation of the follower response as a function of driving style. The relative speed between vehicles was maintained near zero (\(dv \approx 0\)) to minimize confounding effects, allowing spacing to serve as the primary independent variable in the response analysis.

\subsection{Sample Distribution}

A total of 72 LC trials and 80 Respd trials were conducted (see Table \ref{tab:sample_distribution_combined} for a detailed breakdown). Figure \ref{fig:IndependentVariables} represents the sample distribution for the (a) LC Experiment, and (b) Respd Experiments.

\SampleSize
\IndependentVariables

\section{Data Overview}

This section is divided into two parts: (a) LC Experiments and (b) Respd Experiments. Here, we present sample examples and provide a brief description of the measured target vehicle speeds, target gaps, and accepted gaps observed during the experiments.  

\subsection{LC Experiments}

\subsubsection{Sample Examples}

Figure \ref{fig:OverlayTrajectories} presents a representative scenario from the LC experiments corresponding to indicators, where \(DS=1\) and \(DV=0\). In this scenario, the human driver maintained the LC vehicle, positioned in the right lane, parallel to the lead vehicle in the left lane before. Vehicle positions on lane centerline positions are depicted for three important timestamps, before, during, and after LC maneuver. Figure  \ref{fig:Sample Example LC Experiments} subplots the longitudinal position, longitudinal speed, lateral distance, lateral speed, spacing and time gap (spacing between leader-follower pair divided by speed of following vehicle) profiles. The blue vertical line in each subplot marks the time of Auto Mode activation, while the red vertical line indicates the merge completion time, as recorded by the on-site observer and verified through synchronized video logs. In the lateral position plot, the zero line represents the left lane centerline, and the average lateral offset of 3.7 meters corresponds to the right lane centerline.

 At \(t = 24\) seconds, the LC vehicle reached the beginning of the right-turn pavement markings, and the human driver activated Auto Mode. The tAV took over control immediately. At the same moment, the tAV signaled its intent to change lanes by activating the left-turn blinker. The tAV began to decelerate and remained in the subject lane for another 4-5 seconds to perform longitudinal adjustments. Due to its negative relative speed with respect to the lead vehicle, the lead spacing increased during this period. The tAV initiated lateral movement around \(t = 29\) seconds and completed the merge by approximately \(t = 34\) seconds, entering the target lane with a speed difference of about 0.4~m/s relative to the lead vehicle. 

Figure~\ref{fig:OverlayTrajectories 2} and~\ref{fig:Sample Example LC Experiments 2} presents another representative scenario from the LC experiments, corresponding to conditions where \(DS = 2\) and \(DV = 0\). In this scenario, the human driver maintained the lane-change (LC) vehicle in the right lane, positioned close to the first follower (F1). 

From the two examples, we can observe that the tAV employs distinct execution strategies: in one instance, it decelerates before speed matching (Example 1), while in another, it accelerates prior to speed matching (Example 2). These strategies result in varying impacts on the following vehicle. For instance, in Example 1, the F1 vehicle’s speed decreases by 1.5 m/s, whereas in Example 2, the reduction is 2.3 m/s. This differential response highlights the influence of relative speed and spacing on the tAV’s lane-changing (LC) execution strategy which in turn affects follower behavior. The NC-tALC dataset can further be explored to offer critical insights into tAV lane-changing behavior dynamics and their broader impacts.

\OverlayTrajectories
\SampleExampleLC
\OverlayTrajectoriesT
\SampleExampleLCT

\subsubsection{Target Vehicles Speeds and Gaps}

Figure~\ref{fig:Target Vehicles Speed in LC Experiments} illustrates the speed profiles of target vehicles in LC experiments, measured at the time of Auto Mode activation. The data reveal a consistent speed differential between the lead vehicle and the F1, while F1 and F2 exhibit nearly identical speed profiles. At Auto Mode activation, the lead vehicle maintained a higher average speed (17.89 m/s) compared to F1 (17.52 m/s), resulting in a speed difference of approximately 0.37 m/s. In contrast, F1 and F2 displayed closely aligned statistical distributions, with equivalent mean, median, and inter-quartile ranges. Notably, the odometer readings for all target vehicles indicated a speed of 40 mph (~17.88 m/s) at Auto Mode activation. This suggest that lead vehicle displayed speed and GNSS speed is precise, whereas tAVs displayed speed and GNSS speed have difference of ~1 mph. However, the lead vehicle's speed exhibited greater variability across trials. 

Figure~\ref{fig:Target Gap in LC Experiment} depicts the target gap distribution at Auto Mode activation. We expected to have identical target gaps between Lead-F1 and F1-F2. However, the mean first gap (between the lead vehicle and F1) was 36.90 m, while the mean second gap (between F1 and F2) was 29.22 m. This difference in means is likely attributable to the speed differential between the lead vehicle and F1, which is negligible between F1 and F2. 

Additionally, the first gap exhibited a higher standard deviation than the second gap, likely due to greater variability in the lead vehicle's speed across trials compared to that of F1. Another contributing factor we identified is that, in some instances, the follower tAVs required substantial time to reach equilibrium—i.e., attaining a speed of 40 mph from 39 mph. Notably, in 3-4 cases, the target gap varied by approximately 5 meters. One significant outlier was observed when the F1 driver delayed adjusting the set speed, resulting in a considerably larger gap during that trial. Overall, we found that, in the majority of cases, each target gap maintained consistent values within limits, particularly when all vehicles had attained the set speed of 40 mph.

\TargetVehiclesSpeedLC
\TargetGapLC

\subsubsection{Independent Variables and Accepted Gaps}
Figure \ref{fig:Gap Acceptance} (a) illustrates the distribution of merging gaps across varying \(ds-dv\) in LC experiments. The accepted gaps are categorized by the position of the LC vehicle at the Auto Mode activation and their corresponding merging gap: orange dots indicate merges in front of the lead vehicle (Zero), blue dots represent merges in the First gap (between the lead vehicle and the first follower, F1), and green dots denote merges in the Second gap (between F1 and the second follower, F2). In 70\% of cases (51 instances), the tAV merged into the first gap, in 14\% (10 instances) into the second gap, and in 16\% (11 instances) in front of the lead vehicle.

When the relative speed is zero (\(dv = 0\)), the tAV typically merged into the immediate first gap, except in edge-case scenarios. Two instances were observed where the LC vehicle was positioned ahead of the lead vehicle (with \(ds > -5 \, \text{m}\)); in both cases, the tAV merged into the first gap. In all other cases under this condition, the tAV merged in front of the lead vehicle. Additionally, two cases were noted where the LC vehicle was positioned ahead of F1, leading it to reject the immediate first gap and merge into the second gap.

When \(dv\) is negative (tAV slower than the lead vehicle), eight cases involved merging into the first gap, while two cases resulted in merging in front of the lead vehicle, with \(ds\) adjusted to \(-20 \, \text{m}\). Furthermore, in six cases where \(ds > 20 \, \text{m}\), the tAV rejected the first gap and accepted the second gap. This indicates that lower relative speeds shift the acceptable \(ds\) values for the first gap toward lower thresholds, supporting our earlier conjecture that reduced speed moves the overall gap acceptance zone (easy or difficult) to the right (left in Figure~\ref{fig:Gap Acceptance} (a)).

When \(dv\) is positive (tAV faster than the lead vehicle), three cases were recorded where the LC vehicle was behind the lead vehicle, prompting the tAV to merge in front of the lead vehicle. Similarly, three to four cases were observed where the tAV, positioned behind F1, accelerated to merge into the first gap. This suggests that positive relative speeds shift the acceptable \(ds\) values for the first gap toward higher thresholds, consistent with our prior conjecture that increased speed moves the overall gap acceptance zone (easy or difficult) to the left (right in Figure~\ref{fig:Gap Acceptance} (a)).

However, the entire mechanism is highly complex and extends beyond the scope of this paper. These findings contribute valuable data to the NC-tALC dataset, enhancing the understanding of tAV gap acceptance dynamics under varied relative speed and spacing conditions.

\GapAcceptance

\subsection{Respd Experiments}

\subsubsection{Sample Example}

Figure~\ref{fig:Sample Example Respd Experiments} illustrates a sample scenario where the desired spacing (ds) equals 0 m, with the configuration set to HHH, indicating that the LC vehicle and F1 operated in \textit{Hurry} mode. The speed profiles of the follower tAVs demonstrate greater variability compared to the lead vehicle prior to the activation of Auto Mode, a difference likely attributable to the desired speed settings, which were established as the speed limit (45 mph) with an additional speed offset. The lead vehicle maintained a consistent speed of 40 mph, serving as a stable reference.

The LC vehicle initiated Auto Mode at t = 22 s, commencing deceleration within 1 second to adjust its longitudinal profile. After 5-6 seconds later (at t = 28 s), the tAV began its lateral maneuver at a reduced speed, exhibiting a relative speed of 1.8 m/s. During the lane-change process, the tAV engaged in a speed-matching procedure to align with the target lane’s speed. The lane change was successfully completed with a residual speed difference of 1 m/s relative to the lead vehicle.

\SampleExampleRespd

\subsubsection{Target Vehicles Speeds and Gaps}
The lead vehicle demonstrates consistency similar to that observed in LC experiments, maintaining a mean speed of 17.90 m/s with a low standard deviation of 0.09 m/s. In contrast, the follower vehicles (F1 and F2) exhibit greater variability likely due to their tAV driving characteristics. F1 has a mean speed of 18.24 m/s, a standard deviation of 0.43 m/s. F2, also with a mean of 18.24 m/s, shows a higher standard deviation of 0.67 m/s. This indicates that both followers are less stable than the lead vehicle, with F2 being the more variable of the two due to its larger standard deviation and broader range.

The evidence from the box plots in Figure~\ref{fig:Target Vehicles Speed in LC Experiments} shows that F1’s speed variability increases under HHH and CHH configurations, with the upper range extending to 18.93 m/s, while HCC and HCC configurations cluster closer to the mean of 18.24 m/s, reflecting a standard deviation of 0.43 m/s.
F2 exhibits even greater variability, with the box plots indicating wider spreads under HHH and CHH configurations, where the maximum speed reaches 19.84 m/s, supported by a higher standard deviation of 0.67 m/s compared to the tighter clustering under HCC and CCC settings. The speed distribution highlights that under \textit{Hurry} mode, the tAV is exhibiting higher speed variation compared to the \textit{Chill} mode, suggest that \textit{Hurry} mode prefers aggressive driving behavior and \textit{Chill} mode prefer the conservative driving behavior.

\TargetVehiclesSpeedRespd

Figure~\ref{fig:Target Gap in Respd Experiment} displays the target gap distributions in the Respd Experiments. In contrast to the LC experiments, where the mean of the first target gap exceeded that of the second, the Respd Experiments reveal that the second target gap (31.92 m) surpasses the first target gap (28.92 m). This reversal is likely attributable to the higher speed variability observed in the second tAV compared to the first tAV.

\TargetGapRespd

\subsubsection{Independent Variables and Accepted Gaps}

Figure~\ref{fig:Gap Acceptance} (b) illustrates the distribution of merging gaps across varying relative spacings (\(ds\)) and driving style configurations (HHH, HCC, CHH, CCC) for the tAV. When \(ds\) exceeds \(-5 \, \text{m}\) (\(ds > -5 \, \text{m}\)), the tAV merged into the first gap (between the lead vehicle and the first follower, F1) in six cases across the HHH, HCC, and CHH configurations. For \(ds\) less than \(-5 \, \text{m}\) (\(ds < -5 \, \text{m}\)), merging behavior varied by configuration: in four cases under HHH, the tAV merged in front of the lead vehicle; in five cases under CCC and four cases under CHH, the tAV merged into the zero gap (immediately adjacent to the lead vehicle). The data indicate that the tAV merged in front of the lead vehicle, into the first gap, and into the second gap across all driving configurations. Future studies should investigate whether tAV driving configurations influence merging behavior. Notably, the primary objective of this experiment was to assess the response of tAV due to cut-in behavior, and not vice-versa.

\section{Discussion}

\subsection{Use Cases}

Several studies on lane-changing behavior have emphasized the critical need for high-accuracy lane-changing datasets to quantify their impact on traffic flow and safety \cite{Ali2025EmpiricalNeeds, Liu2023VehicleReview,Ma2023AResearch, Wang2019ReviewAnalyses}. The high-accuracy trajectory dataset presented in this study, collected using RTK-GPS devices, facilitates a deeper understanding of how AVs can influence traffic dynamics and safety outcomes during lane-changing maneuvers. Given the limited sample size, the dataset serves as a valuable resource as expert data for lane change decision models, trajectory prediction, and execution frameworks, providing precise data for algorithm validation. Further, This dataset can support the development of behavioral models for AVs, enabling researchers to simulate and predict interaction patterns and responses in mixed traffic environments.  Furthermore, it enables comparative analyses of lane-changing behaviors between AVs and human drivers, shedding light on differences in decision-making and operational efficiency across vehicle classes. By offering a foundation for these applications, the dataset contributes to the design of safer and more efficient autonomous driving systems, with potential extensions to diverse traffic scenarios and vehicle types.

\subsection{Challenges}
The data collection effort for this study required multiple iterations at different sites to identify an optimal site that balanced the need for high-quality, efficient data capture with minimal disruption to live traffic. As tAVs offer unique and new driving experience, extensive training was provided to all participants before experiments, equipping them with the skills to navigate experimental scenarios safely while prioritizing the well-being of themselves and surrounding road users. 

However, several challenges emerged during the process. A significant bottleneck was the frequent interruption by other road users before, during and after the experiment trials. Interruptions disrupted the controlled conditions and coordination among drivers to complext the experiment scenario, which necessitate repetition. Additionally, technical malfunctions posed substantial hurdles, including intermittent internet connectivity losses, INS gyroscope failures, RTK signal degradation due to environmental factors such as urban obstructions, and battery depletion in data collection equipment. These issues necessitated robust contingency planning, such as redundant power supplies and backup communication systems, to maintain data integrity. Addressing these challenges required meticulous coordination and adaptive strategies to ensure the reliability and validity of the collected trajectory dataset.

\subsection{Limitation}

A key limitation of the dataset presented in this study is its relatively small sample size, comprising 72 + 80 cases, which will need detailed and careful investigation  to generalize the findings for research applications. The experiments were conducted in a simplified setup, utilizing a single speed level and a single type of autonomous vehicle (tAV), which limits the dataset’s ability to capture the full spectrum of real-world AV-AV interaction scenarios. These constraints may affect the applicability of the dataset to more complex traffic environments or diverse vehicle configurations. However, the experimental framework is designed to be scalable, allowing for future expansions to include varied speed profiles, multiple tAV types, and diverse environmental conditions. Such extensions could enhance the dataset’s robustness and broaden its utility for studying a wider range of autonomous vehicle behaviors and their impacts on traffic flow and safety.

\section{Conclusion}

This study presents a high-accuracy trajectory dataset to examine lane-changing and cut-in responding behavior of tAV, and quantify its impact of traffic flow and safety. The experiments were conducted  on test site requiring mandatory lane change under controlled settings. The first set of experiments were designed to study the impact of the relative speed and relative spacing on the tAV's lane-changing behavior, including decision-making, execution and impact on target following vehicles. The second set of experiments were designed to study the cut-in response of the tAV, when tAV have different driving style (conservative or aggressive). 

Additionally, this dataset is helpful in understanding automated vehicle-automated vehicle interaction behaviors in empirical settings as all vehicle were driven under ACC / Auto Mode settings. The detailed description of the data collection setup ensures clarity and supports reproducibility. The experimental framework can be extended to explore varied speeds, locations, and AV types in future work.  This work provides a practical resource for researchers studying autonomous vehicle interactions, with opportunities for further development to enhance AV system reliability.

\section{AUTHOR CONTRIBUTIONS}

Abhinav Sharma: Conceptualization, Methodology, Data collection, analysis, Writing - original draft. Zijun He: Conceptualization, Data collection. Danjue Chen: Conceptualization, Methodology, Supervision, Funding acquisition. 

\section{DECLARATION OF CONFLICTING INTERESTS}

The authors declared no potential conflicts of interest with respect to the research, authorship, and/or publication of this article.

\section{ACKNOWLEDGMENTS}

This research has been funded by National Science Foundation (NSF) Award \(\#\ 535015\).

\newpage

\bibliographystyle{trb}
\bibliography{References}

@misc{Ma2023AResearch,
    title = {{A review of vehicle lane change research}},
    year = {2023},
    booktitle = {Physica A: Statistical Mechanics and its Applications},
    author = {Ma, Changxi and Li, Dong},
    month = {9},
    volume = {626},
    publisher = {Elsevier B.V.},
    doi = {10.1016/j.physa.2023.129060},
    issn = {03784371}
}

@misc{Apnews2025Apnews,
    title = {{Apnews}},
    year = {2025},
    author = {{Apnews}},
    url = {https://apnews.com/article/tesla-musk-self-driving-analyst-automated-traffic-a4cc507d36bd28b6428143fea80278ce#:~:text=Full%20Self%2DDriving%20is%20being,more%20for%20the%20optional%20system.}
}

@article{Ali2025AutonomousFollower,
    title = {{Autonomous vehicle lane-changing dynamics and impact on the immediate follower}},
    year = {2025},
    journal = {Analytic Methods in Accident Research},
    author = {Ali, Yasir},
    month = {6},
    volume = {46},
    publisher = {Elsevier Ltd},
    doi = {10.1016/j.amar.2025.100388},
    issn = {22136657}
}

@misc{Buyacar2025Buyacar,
    title = {{Buyacar}},
    year = {2025},
    author = {{Buyacar}},
    url = {https://www.buyacar.co.uk/the-latest-tesla-statistics/#:~:text=How%20many%20Teslas%20are%20there,around%20400%2C000%20cars%20each%20quarter.}
}

@article{Ali2018ConnectivitysStudy,
    title = {{Connectivity's impact on mandatory lane-changing behaviour: Evidences from a driving simulator study}},
    year = {2018},
    journal = {Transportation Research Part C: Emerging Technologies},
    author = {Ali, Yasir and Zheng, Zuduo and Haque, Md Mazharul},
    month = {8},
    pages = {292--309},
    volume = {93},
    publisher = {Elsevier Ltd},
    doi = {10.1016/j.trc.2018.06.008},
    issn = {0968090X}
}

@misc{Ali2025EmpiricalNeeds,
    title = {{Empirical research on car-following and lane-changing: Recent developments, emerging vehicle technologies’ impact, and future research needs}},
    year = {2025},
    booktitle = {Transportation Research Interdisciplinary Perspectives},
    author = {Ali, Yasir and Sharma, Anshuman and Zheng, Zuduo},
    month = {5},
    volume = {31},
    publisher = {Elsevier Ltd},
    doi = {10.1016/j.trip.2025.101368},
    issn = {25901982}
}

@article{Shi2021EmpiricalSettings,
    title = {{Empirical study on car-following characteristics of commercial automated vehicles with different headway settings}},
    year = {2021},
    journal = {Transportation Research Part C: Emerging Technologies},
    author = {Shi, Xiaowei and Li, Xiaopeng},
    month = {7},
    volume = {128},
    publisher = {Elsevier Ltd},
    doi = {10.1016/j.trc.2021.103134},
    issn = {0968090X}
}

@article{Ammourah2024IntroductionExtraction,
    title = {{Introduction to the Third Generation Simulation Dataset: Data Collection and Trajectory Extraction}},
    year = {2024},
    journal = {Transportation Research Record: Journal of the Transportation Research Board},
    author = {Ammourah, Rami and Beigi, Pedram and Fan, Bingyi and Hamdar, Samer H. and Hourdos, John and Hsiao, Chun-Chien and James, Rachel and Khajeh-Hosseini, Mohammdreza and Mahmassani, Hani S. and Monzer, Dana and Radvand, Tina and Talebpour, Alireza and Yousefi, Mahdi and Zhang, Yanlin},
    month = {7},
    doi = {10.1177/03611981241257257},
    issn = {0361-1981}
}

@article{Ali2024InvestigatingDataset,
    title = {{Investigating autonomous vehicle discretionary lane-changing execution behaviour: Similarities, differences, and insights from Waymo dataset}},
    year = {2024},
    journal = {Analytic Methods in Accident Research},
    author = {Ali, Yasir and Sharma, Anshuman and Chen, Danjue},
    month = {6},
    volume = {42},
    publisher = {Elsevier Ltd},
    doi = {10.1016/j.amar.2024.100332},
    issn = {22136657}
}

@misc{FHWA2006NGSIM,
    title = {{NGSIM }},
    year = {2006},
    booktitle = {2006},
    author = {{FHWA}},
    url = {https://ops.fhwa.dot.gov/trafficanalysistools/ngsim.htm}
}

@article{Zheng2014RecentChanging,
    title = {{Recent developments and research needs in modeling lane changing}},
    year = {2014},
    journal = {Transportation Research Part B: Methodological},
    author = {Zheng, Zuduo},
    pages = {16--32},
    volume = {60},
    publisher = {Elsevier Ltd},
    doi = {10.1016/j.trb.2013.11.009},
    issn = {01912615}
}

@misc{Wang2019ReviewAnalyses,
    title = {{Review of Lane-Changing Maneuvers of Connected and Automated Vehicles: Models, Algorithms and Traffic Impact Analyses}},
    year = {2019},
    booktitle = {Journal of the Indian Institute of Science},
    author = {Wang, Zhen and Shi, Xiaowei and Li, Xiaopeng},
    number = {4},
    month = {12},
    pages = {589--599},
    volume = {99},
    publisher = {Springer},
    doi = {10.1007/s41745-019-00127-7},
    issn = {00194964}
}

@misc{SuperCruise2025SuperCruise,
    title = {{Super Cruise}},
    year = {2025},
    author = {{Super Cruise}},
    url = {https://www.theverge.com/2021/7/23/22589285/gm-super-cruise-automatic-lane-change-gmc-chevy-silverado}
}

@misc{Tesla2025TeslaAutopilot,
    title = {{Tesla Autopilot}},
    year = {2025},
    author = {{Tesla}},
    month = {1},
    url = {https://www.tesla.com/support/autopilot}
}

@article{Liu2023VehicleReview,
    title = {{Vehicle Lane Change Models—A Historical Review}},
    year = {2023},
    journal = {Applied Sciences},
    author = {Liu, Xinchao and Hong, Liang and Lin, Yier},
    number = {22},
    month = {11},
    pages = {12366},
    volume = {13},
    publisher = {MDPI AG},
    doi = {10.3390/app132212366}
}
\end{document}